\pdfoutput=1
\documentclass[11pt]{article}

\usepackage[final]{acl}

\usepackage{times}
\usepackage{latexsym}
\usepackage{amsmath}
\usepackage{enumitem}
\usepackage{makecell}
\usepackage{multirow}
\usepackage{amsfonts}
\usepackage{booktabs}
\usepackage{tabularx}
\usepackage{array}
\usepackage{cleveref}
\usepackage{tcolorbox}
\usepackage{float}
\tcbuselibrary{breakable}
\usepackage[T1]{fontenc}
\usepackage[table]{xcolor}  % 在导言区加载
\definecolor{myblue}{RGB}{230, 242, 255}  % 放在导言区
\definecolor{tableheadcolor}{gray}{0.92} % 表头浅灰色
\definecolor{tablerowcolor}{gray}{0.96} % 偶数行极浅灰色
\definecolor{scorezero}{RGB}{190,35,45}
\definecolor{scoreone}{RGB}{205,100,20}
\definecolor{scoretwo}{RGB}{170,125,0}
\definecolor{scorethree}{RGB}{45,140,75}
\definecolor{scorefour}{RGB}{0,95,50}
\tcbset{appendixtext/.style={colback=white,colframe=black!35,width=\textwidth,boxrule=0.4pt,arc=1pt,boxsep=0pt,left=5pt,right=5pt,top=4pt,bottom=4pt,before upper={\setlength{\parindent}{0pt}\setlength{\parskip}{1pt}}}}
\newcommand{\appendixheading}[1]{\par\noindent{\setlength{\fboxsep}{3pt}\colorbox{gray!12}{\parbox{\dimexpr\linewidth-2\fboxsep\relax}{\strut\textbf{#1}\strut}}}\par\vspace{2pt}}
\newcommand{\appendixfield}[1]{\par\addvspace{2pt}#1\par}

\newcommand{\kailabel}[1]{{\CJKfamily{gkai}#1}}
\newcommand{\scorezero}[1]{\textcolor{scorezero}{#1}}
\newcommand{\scoreone}[1]{\textcolor{scoreone}{#1}}
\newcommand{\scoretwo}[1]{\textcolor{scoretwo}{#1}}
\newcommand{\scorethree}[1]{\textcolor{scorethree}{#1}}
\newcommand{\scorefour}[1]{\textcolor{scorefour}{#1}}
\newcommand{\cell}[4]{#1{\scriptsize\hspace{1pt}$\pm$#2}\hspace{3pt}{\scriptsize(#3\,/\,#4)}}
\newcommand{\avg}[3]{\textit{#1}{\scriptsize\phantom{\hspace{1pt}$\pm$0.00}}\hspace{3pt}{\scriptsize(\textit{#2}\,/\,\textit{#3})}}
\usepackage[utf8]{inputenc}
\usepackage{CJKutf8}
\usepackage{microtype}
\usepackage{graphicx}

\usepackage{inconsolata}
\usepackage{graphicx}
\usepackage{subcaption}

\title{\texttt{APRiL}: An Open-ended Benchmark for Chinese Social Awareness in Large Language Models}

\author{
Tian Lan$^{1}$,
Han Zhang$^{2}$,
Shanshan Wang$^{3}$, 
Yahan Yu$^{4}$,
Jiang Li$^{5}$,
Yugo Murawaki$^{1}$\thanks{\ \ Corresponding Author}
 \\
$^1$Kyoto University, 
$^2$Shanghai Jiao Tong University,\\
$^3$University of Macau,
$^4$The University of Tokyo,
$^5$Inner Mongolia University
\\
\texttt{velikayascarlet@gmail.com}}

\usepackage{fontawesome5}

\begin{document}
\maketitle
\begin{abstract}

Large language models are increasingly used in social interactions, where appropriate responses require recognizing implicit cues, inferring mental states, identifying conflicts, and considering potential consequences. However, existing benchmarks often focus on specific social abilities, constrained task formats, or particular interaction settings, offering limited coverage of social awareness across diverse contexts. To address this gap, we introduce APRiL, an open-ended benchmark for Chinese social awareness with instance-specific rubrics. APRiL contains 2,196 Social Awareness Evaluation Instances across six categories and two complementary tasks: Social Scenario Analysis evaluates models’ understanding of social cues, mental states, and conflict structures, while Socially Aware Response assesses whether they can translate this understanding into appropriate, natural responses. Experiments on 12 models from six families reveal a consistent understanding-to-response gap. Higher-quality scenario analyses improve responses, while weak analyses may reduce response quality. APRiL provides a systematic resource for advancing socially aware language models in Chinese contexts. Our data and code are available at \href{https://anonymous.4open.science/r/APRiL-05DE/}{APRiL}.

\end{abstract}

\begin{CJK}{UTF8}{gbsn}

\section{Introduction}
\begin{figure}[t]
\centering
\includegraphics[width=\columnwidth]{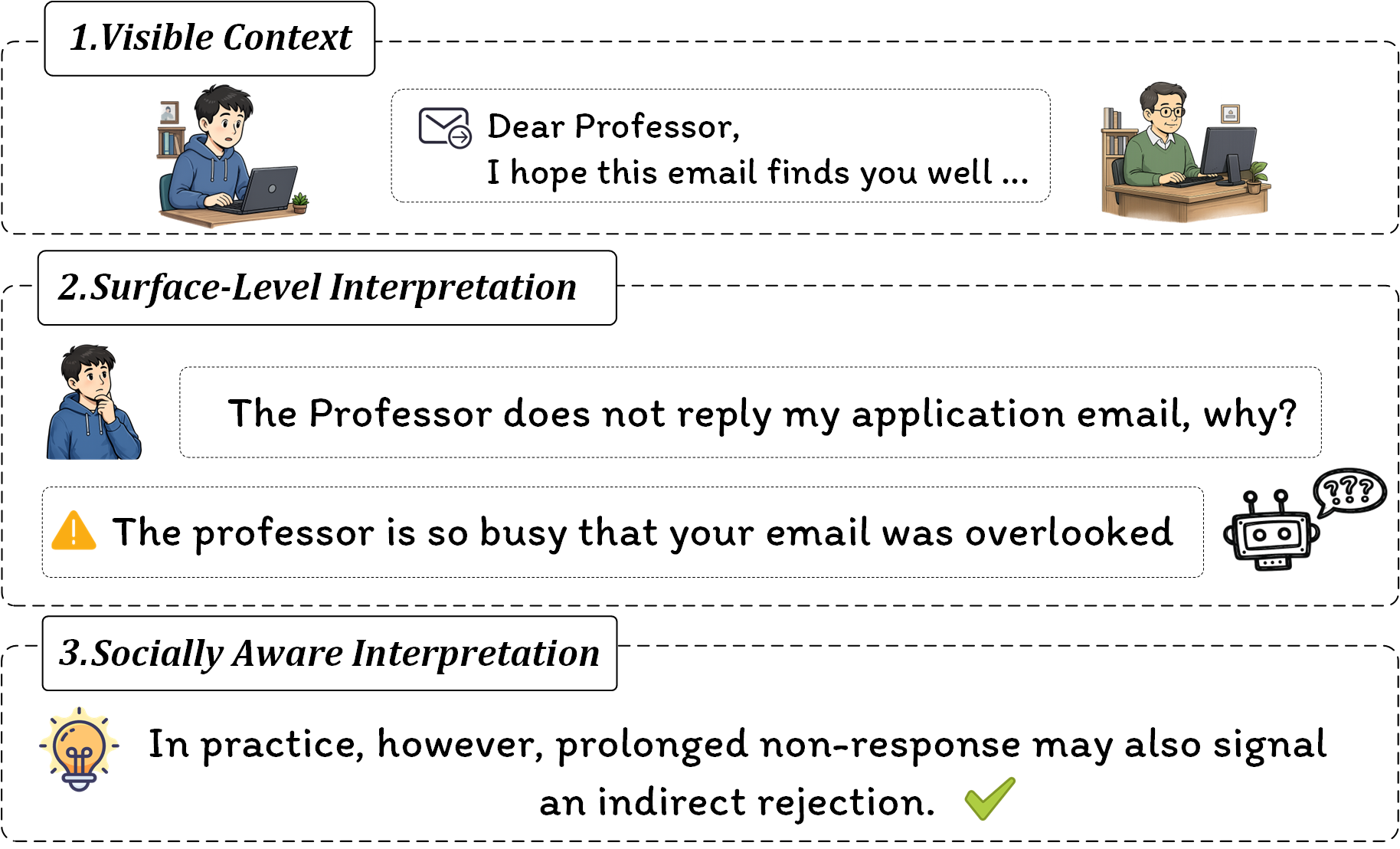}
\caption{An example of implicit meaning in social interaction.}
\label{fig:social-example}
\end{figure}

\begin{figure*}[t]
\centering
\includegraphics[width=\textwidth]{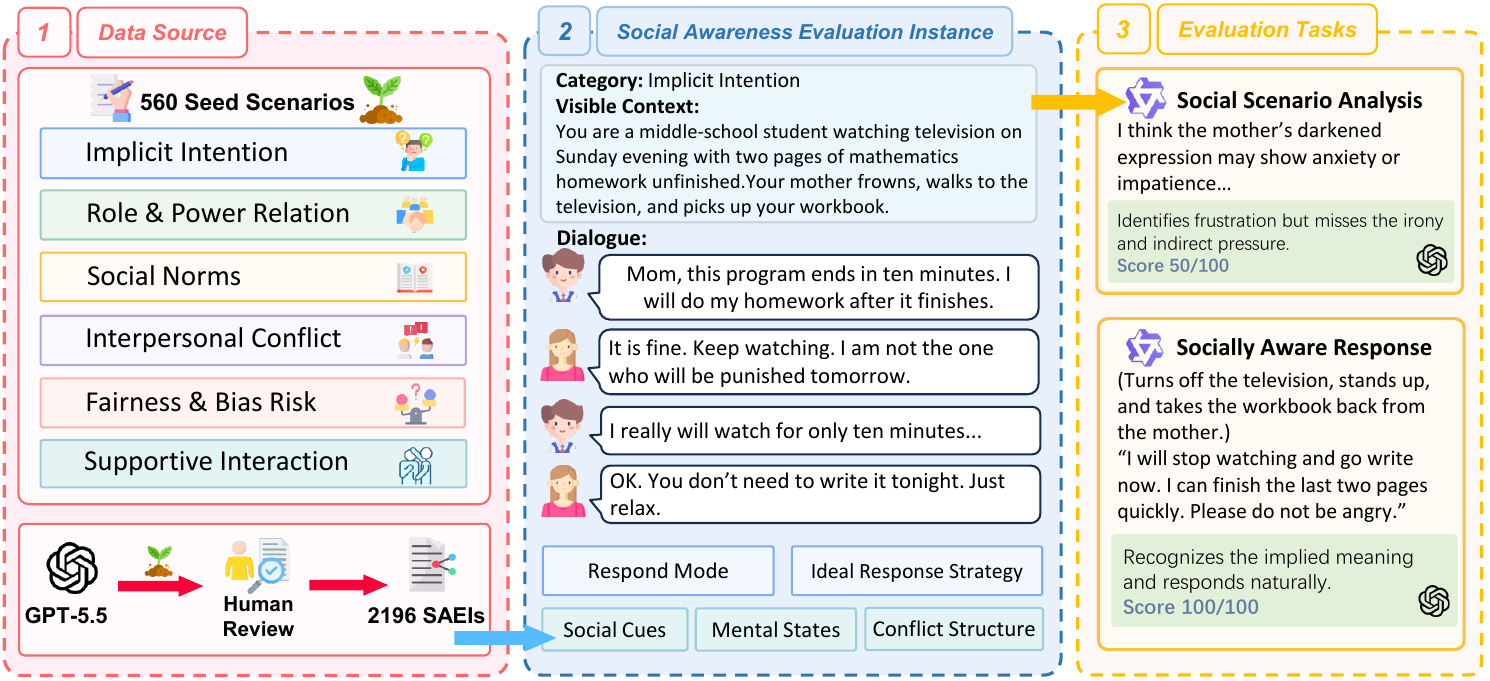}
\caption{Overview of APRiL’s dataset construction, SAEI structure, and evaluation tasks.}
\label{fig:overall}
\end{figure*}
Large language models (LLMs) have achieved strong performance across a wide range of natural language processing (NLP) tasks~\citep{wang-etal-2024-best,li-etal-2026-wrote,lin-etal-2026-exploring,lan-etal-2026-cedar} and are increasingly deployed in social interactions such as cross-cultural communication~\citep{li2024culturepark}, business communication~\citep{subrahmanyam2026collaboration}, emotional support~\citep{zheng2025customizing}, and team conflict mediation~\citep{westermann2023llmediator}. Unlike conventional question answering, an appropriate response in social interaction depends not only on textual content, but also on interpersonal relationships, mental states, power differences, social norms, and potential consequences. Even when a model correctly understands the surface meaning of an utterance, neglecting these factors may lead it to make unwarranted attributions, push an interaction too far, or respond in a manner inconsistent with the relationship.

Reliable and adaptive communication in real-world social interaction therefore requires LLMs to possess social awareness. In this paper, we define social awareness as the ability of a model to identify explicit and implicit cues in social interactions, understand participants' relational structures, mental states, and potential conflicts, and use this information to generate responses that fit the interpersonal relationship and interaction setting while accounting for real-world consequences~\citep{yang-etal-2025-socially,premack1978does,ohbuchi1997multiple}.

Crucial information in social interaction is often conveyed through indirect expressions, silence, changes in tone, and differences between public and private settings. This is especially common in Chinese, where politeness norms~\citep{gu1990politeness} and the role of face~\citep{mao1994beyond} shape how intentions are expressed. The same utterance may also convey different meanings across relationships and situations, as illustrated in \textbf{Appendix~\ref{app:dataset-details}}. As shown in Figure~\ref{fig:social-example}, when an applicant receives no response from a professor for an extended period, an LLM may attribute the silence merely to the professor's busy schedule. In practice, however, prolonged non-response may also signal an indirect rejection. A model that fails to recognize this possibility may advise the applicant to keep following up, thereby placing additional pressure on the professor and exposing the applicant to interpersonal risk. As LLMs increasingly enter interaction-sensitive domains such as education, business, and healthcare, accurately understanding such social information has become particularly important.

Although prior work has examined the social capabilities of LLMs, existing evaluations usually focus on particular subskills. Social commonsense reasoning benchmarks~\citep{sap-etal-2019-social,forbes-etal-2020-social} primarily assess knowledge of social norms and reasons for behavior; Theory-of-Mind benchmarks~\citep{chen-etal-2024-tombench,xu-etal-2024-opentom} focus on reasoning about mental states; and evaluations of morality~\citep{yu-etal-2024-cmoraleval}, fairness~\citep{lan-etal-2025-mcbe,lan-etal-2025-f2bench}, and safety~\citep{zhang-etal-2024-safetybench,mazeika2024harmbench} respectively examine normative judgment, bias risk, and harmful behavior. However, these benchmarks rarely test whether a model can jointly interpret social cues, mental states, and conflict structures under incomplete information, asymmetric relationships, and competing goals, and then turn this understanding into an appropriate response. In addition, many existing evaluations are built around a narrow set of scenario types or interaction formats, such as classification items~\citep{zhan2023socialdial}, predefined options~\citep{zhou-etal-2025-socialeval}, or a single platform~\citep{huang2025si}. As a result, they provide limited evidence of whether a model's social awareness holds across diverse relationships and interaction settings.

To address this gap, we propose APRiL, \textbf{A}n O\textbf{P}en-ended Benchmark for Social Awa\textbf{R}eness \textbf{i}n \textbf{L}arge Language Models. APRiL systematically evaluates LLM social awareness in Chinese contexts. As illustrated in Figure~\ref{fig:overall}, APRiL contains 2,196 Social Awareness Evaluation Instances (SAEIs) across six representative categories: Implicit Intention, Role and Power Relation, Social Norms, Interpersonal Conflict, Fairness and Bias Risk, and Supportive Interaction. We design two complementary tasks, Social Scenario Analysis (SSA) and Socially Aware Response (SAR). SSA assesses whether models can identify social cues and understand participants' mental states and conflict structures. SAR assesses whether they can transform this understanding into natural and appropriate responses, accounting for interaction settings and real-world consequences. We further conduct a systematic evaluation of 12 models from six mainstream model families.

Our contributions are summarized as follows:

\begin{itemize}[leftmargin=*]
\item \textbf{Social Awareness Perspective.} We study LLMs' performance in Chinese social contexts from the perspective of social awareness, focusing on whether models can integrate implicit social information and translate their understanding into appropriate responses.
\item \textbf{Evaluation Benchmark.} We introduce APRiL, comprising 2,196 SAEIs across six categories, with two open-ended tasks and instance-specific scoring rubrics.
\item \textbf{Experimental Analysis.} We evaluate 12 models from six families and analyze the gap between scenario understanding and response generation, as well as how providing scenario analyses of varying quality affects responses.
\end{itemize}

\section{Related Works}
\subsection{Social Awareness and Cognitive Foundations}
Social awareness in language technology concerns the social information needed to interpret and act within an interaction~\citep{yang-etal-2025-socially}. Social information-processing research connects the interpretation of social cues with interaction goals, response construction, and the evaluation of consequences~\citep{crick1994review}. Theory of Mind provides a basis for inferring others' knowledge, beliefs, and intentions~\citep{premack1978does}, while interpersonal-conflict research highlights competing task, relationship, identity, fairness, and power goals~\citep{ohbuchi1997multiple}. Together, these studies motivate APRiL's attention to observable cues, evidence-grounded mental-state inference, latent conflicts, and the response that follows.

\subsection{Social Understanding and Interaction in LLMs}

Existing benchmarks evaluate different aspects of social understanding. Social IQa~\citep{sap-etal-2019-social} examines the causes and consequences of social behavior, while Social Chemistry 101~\citep{forbes-etal-2020-social} and NormBank~\citep{ziems-etal-2023-normbank} focus on social norms and their contextual variation. ToMBench~\citep{chen-etal-2024-tombench} and OpenToM~\citep{xu-etal-2024-opentom} evaluate reasoning about participants' mental states. PUB~\citep{sravanthi-etal-2024-pub} and MultiPragEval~\citep{park-etal-2024-multiprageval} examine pragmatic and implicit meaning, while EQ-Bench~\citep{paech2023eq} evaluates emotional understanding in English conflict dialogues. APRiL evaluates these aspects within open-ended social scenarios and examines both scenario understanding and response generation.

Several benchmarks evaluate social abilities through dialogue and interactive tasks. SocialDial~\citep{zhan2023socialdial} uses Chinese multi-turn dialogues for social-factor classification and norm-violation detection. SOTOPIA~\citep{zhou2024sotopia} evaluates agents through continuous interactions aimed at achieving social goals, such as persuading a partner to change a behavior or negotiating a sale, while SocialEval~\citep{zhou-etal-2025-socialeval} uses branching narratives in which models select predefined responses and are evaluated according to goal achievement and interpersonal abilities. ToMELP~\citep{wang-etal-2026-tomelp} focuses on using mental-state inference to generate controllable responses in persuasive dialogue. SI-Bench~\citep{huang2025si} focuses on social-platform conversations, mainly examining social relationships, relationship development, and interaction strategies in dyadic dialogues. 

Unlike SOTOPIA's simulated goal-driven interactions and SI-Bench's social-platform dialogues, APRiL covers six representative categories spanning a wider variety of Chinese social contexts. It separately evaluates fine-grained scenario understanding and open-ended response generation through Social Scenario Analysis and Socially Aware Response.

\section{The Dataset}
APRiL contains 2,196 Social Awareness Evaluation Instances (SAEIs). The evaluated model sees only the Visible Context and Dialogue; the remaining annotations guide rubric construction and judging. The attributes and a bilingual example are provided in \textbf{Appendix~\ref{app:dataset-details}}.

\begin{table*}[t]
\centering
\small
\rowcolors{2}{white}{tablerowcolor}
\resizebox{\textwidth}{!}{%
\begin{tabular}{lrrrr}
\toprule
\rowcolor{tableheadcolor} \textbf{Category} & \textbf{Instances} & \textbf{Proportion} & \textbf{Avg. Dialogue Turns} & \textbf{Avg. Dialogue Length} \\
\midrule
Implicit Intention & 382 & 17.40\% & 7.46 & 298.33 \\
Role and Power Relation & 447 & 20.36\% & 6.83 & 308.26 \\
Social Norms & 339 & 15.44\% & 6.78 & 245.32 \\
Interpersonal Conflict & 318 & 14.48\% & 7.28 & 343.15 \\
Fairness and Bias Risk & 391 & 17.81\% & 7.34 & 393.76 \\
Supportive Interaction & 319 & 14.53\% & 8.40 & 320.66 \\
\textbf{Total} & \textbf{2,196} & \textbf{100\%} & \textbf{7.32} & \textbf{318.89} \\
\bottomrule
\end{tabular}%
}
\caption{Statistics of the APRiL benchmark.}
\label{tab:statistics}
\end{table*}
\subsection{Dataset Construction}
APRiL follows a seed-based construction process. We focus on intentions, relationships, norms, conflicts, fairness, and support needs that affect scenario understanding and response appropriateness. With expert guidance, we organize scenarios into six categories: \textbf{Implicit Intention, Role and Power Relation, Social Norms, Interpersonal Conflict, Fairness and Bias Risk, and Supportive Interaction}. These categories are also supported by prior work: SI-Bench~\citep{huang2025si} addresses implicit meaning and power dynamics; SocialDial~\citep{zhan2023socialdial} and SOTOPIA~\citep{zhou2024sotopia} examine social norms; SocialEval~\citep{zhou-etal-2025-socialeval} considers interests and conflicts; McBE~\citep{lan-etal-2025-mcbe} evaluates social bias; and ESConv~\citep{liu-etal-2021-towards} studies emotional support.

We manually designed seeds across the six categories, varying interpersonal relationships, interaction settings, and practical tasks, and retained 560 after review. Each seed includes its central mechanism, surface manifestation, possible meaning, primary tension, and likely model failures. Category definitions, boundaries, and a seed example are provided in \textbf{Appendix~\ref{app:dataset-details}}.

GPT-5.5~\citep{singh2025openai} expanded each seed into multiple SAEIs. Candidates underwent programmatic structural checks and rule-based LLM-assisted quality checks, followed by review from eight native Chinese-speaking graduate-level reviewers. Review focused on cue implicitness, interaction complexity, functional boundaries among hidden annotations, and response constraints. Reviewers checked all annotations against the scenario and dialogue; unsupported, unclear, or inconsistent annotations were rewritten. Two reviewers examined each candidate, resolving disagreements about retention and revision through discussion. Instances with obvious cues, overly short dialogues, unnatural language, or insufficient real-world interaction constraints were removed. The final dataset contains 2,196 reviewed and revised SAEIs.

Each SAEI's rubric was grounded in its hidden annotations and ideal response strategy. Anonymized high- and low-scoring responses from preliminary model runs helped identify alternative valid interpretations and recurring errors. Score points represent independently assessable expected content, while deduction points capture unsupported inferences and interactional risks. Sixteen annotators reviewed the rubrics, merging semantically overlapping points and refining unclear criteria.

\subsection{Statistics}
Table~\ref{tab:statistics} summarizes APRiL, which contains 2,196 SAEIs across six relatively balanced categories, each accounting for 14.48--20.36\% of the dataset. Role and Power Relation is the largest category (20.36\%), while Interpersonal Conflict is the smallest (14.48\%), reducing the risk that aggregate evaluation is dominated by a single category. Each SAEI contains an average of 7.32 dialogue turns and 318.89 Chinese characters. Category-level averages range from 6.78 to 8.40 turns, with Supportive Interaction containing the most turns. Fairness and Bias Risk has the longest average dialogue, whereas Social Norms has the shortest. 

Each SAEI also includes an instance-specific scoring rubric. On average, an instance contains 12.58 score points and 5.65 deduction points for SSA, and 4.91 score points and 4.04 deduction points for SAR.

\section{Evaluation Tasks}
In this paper, social awareness is defined as the ability of a model, in interactions involving other individuals or groups, to identify explicit and implicit social cues, understand participants' relational structures, mental states, and potential conflicts, and transform this information into responses that fit the interpersonal relationship and setting while accounting for real-world consequences~\citep{yang-etal-2025-socially}. APRiL evaluates these related abilities through two complementary tasks: Social Scenario Analysis (SSA) and Socially Aware Response (SAR), described below. Each SAEI includes task-specific score and deduction points, and results are reported on a 0--100 scale.

\subsection{Social Scenario Analysis}
Given the Visible Context and Dialogue, SSA evaluates a model's ability to identify key social information and analyze its effects on participants' understanding and subsequent interaction. SSA has three dimensions. Social Cues measures recognition of observable information such as relationships, verbal behavior, setting differences, and contrasts across interactions. Mental States measures evidence-based inference about participants' information states, intentions, goals, and concerns. Conflict Structure measures understanding of cognitive differences, competing interests, role responsibilities, and value trade-offs.

Scoring also considers coverage of key information, the distinction between fact and inference, and appropriate handling of uncertainty. Merely restating surface content, offering generic analysis, or asserting motives without evidence lowers the score. Alternative interpretations can receive high scores when they are supported by the scenario and fit its relationships and context.

For a set of evaluated instances, the score for each dimension $d$ is computed by pooling matched score points and triggered deduction points within that dimension:

\begin{equation}
S_d =
100 \times
\max\left\{0,
\frac{\sum_i m_{i,d}-\sum_i p_{i,d}}
{\sum_i n_{i,d}}
\right\},
\end{equation}

\noindent where $d\in\{\mathrm{cue},\mathrm{mental},\mathrm{conflict}\}$, and $m_{i,d}$, $p_{i,d}$, and $n_{i,d}$ denote the numbers of matched score points, triggered deduction points, and available score points for instance $i$ in dimension $d$, respectively. The overall SSA score $S_{\mathrm{SSA}}$ is the arithmetic mean of $S_{\mathrm{cue}}$, $S_{\mathrm{mental}}$, and $S_{\mathrm{conflict}}$.

\subsection{Socially Aware Response}
\label{sec:sar}
Socially Aware Response evaluates whether a model can generate a natural and appropriate response to a given social situation, consistent with the current speaker's identity and setting.

SAR assesses whether the response fits the relationship, power structure, setting, and practical task, while considering participants' knowledge, goals, and concerns and avoiding unsupported claims about their motives. When perspectives, interests, or role responsibilities conflict, an appropriate response should consider likely consequences and address the tension through clarification, options, or boundary setting.

Brevity is also part of SAR. Grice's conversational maxims emphasize providing relevant information and avoiding unnecessary verbosity~\citep{grice1975logic}, while user research reports a preference for concise responses and concerns about verbosity~\citep{liu-etal-2025-understand}. To account for communicative burden, a response longer than its Visible Context and Dialogue combined incurs one additional deduction-point trigger. SAR is scored as a single dimension using the same pooled formula.

% SAR uses the same pooled scoring as SSA, aggregated over instances, so the score is a net rate over available score points: each triggered deduction point cancels one matched score point. Because SAR defines more deduction points per score point than SSA ($0.81$ vs.\ $0.44$), the two tasks' absolute score levels are not directly comparable.
% 需要的宏包：booktabs, xcolor[table]
% 需要的宏包：booktabs, graphicx, xcolor[table]

\begin{table*}[t]
\centering
\footnotesize
\setlength{\tabcolsep}{4pt}
\renewcommand{\arraystretch}{0.94}
\rowcolors{2}{white}{blue!5}
\resizebox{\textwidth}{!}{%
\begin{tabular}{l@{\hskip 10pt}c@{\hskip 8pt}c@{\hskip 8pt}c@{\hskip 14pt}c@{\hskip 14pt}c}
\toprule
\rowcolor{blue!15}
\textbf{Model}
 & \shortstack{\textbf{Social}\\\textbf{Cues}}
 & \shortstack{\textbf{Mental}\\\textbf{States}}
 & \shortstack{\textbf{Conflict}\\\textbf{Structure}}
 & \shortstack{\textbf{SSA}\\{\scriptsize($\pm$SD; Match$\uparrow$/Deduct$\downarrow$ rate, \%)}}
 & \shortstack{\textbf{SAR}\\{\scriptsize($\pm$SD; Match$\uparrow$/Deduct$\downarrow$ rate, \%)}} \\
\midrule
\rowcolor{gray!20}
\multicolumn{6}{l}{\textit{Black-box models}} \\
GPT-5.1 & \textbf{84.89} & \underline{76.27} & 86.56 & \cell{\underline{82.57}}{0.21}{85.61}{\underline{6.65}} & \cell{37.65}{0.12}{\textbf{65.89}}{23.44} \\
DS-V4-Pro (T) & \underline{84.09} & \textbf{78.40} & \textbf{88.29} & \cell{\textbf{83.59}}{0.05}{\textbf{87.94}}{9.78} & \cell{33.46}{0.05}{56.06}{26.16} \\
DS-V4-Pro (NT) & 82.91 & 73.89 & \underline{87.52} & \cell{81.44}{0.08}{\underline{87.60}}{13.77} & \cell{25.20}{0.05}{46.80}{25.87} \\
Kimi-K2.5 (T) & 79.75 & 60.56 & 84.44 & \cell{74.92}{0.16}{82.23}{16.08} & \cell{41.39}{0.33}{61.02}{22.71} \\
DS-V4-Flash (T) & 76.40 & 64.07 & 76.71 & \cell{72.39}{0.09}{75.14}{\textbf{5.83}} & \cell{36.52}{0.07}{55.34}{\underline{22.41}} \\
Gemini-3-Pro & 79.84 & 67.55 & 83.58 & \cell{76.99}{0.18}{82.35}{12.00} & \cell{\underline{42.24}}{0.20}{\underline{63.80}}{23.45} \\
Gemini-3.8-Flash & 80.78 & 57.94 & 84.75 & \cell{74.49}{0.13}{81.92}{16.23} & \cell{\textbf{42.88}}{0.12}{59.13}{\textbf{19.18}} \\
Kimi-K2.5 (NT) & 73.89 & 58.16 & 75.87 & \cell{69.31}{0.16}{75.35}{13.19} & \cell{34.34}{0.07}{54.27}{23.52} \\
DS-V4-Flash (NT) & 75.67 & 58.89 & 74.15 & \cell{69.57}{0.08}{73.55}{8.37} & \cell{30.41}{0.04}{51.61}{25.03} \\
\midrule
\rowcolor{gray!20}
\multicolumn{6}{l}{\textit{White-box models}} \\
Ministral-14B & \textbf{83.11} & \textbf{68.71} & \textbf{85.75} & \cell{\textbf{79.19}}{0.07}{\textbf{84.59}}{11.84} & \cell{\textbf{23.15}}{0.12}{\textbf{49.37}}{\underline{28.91}} \\
Ministral-8B & \underline{81.42} & \underline{65.73} & \underline{83.08} & \cell{\underline{76.74}}{0.18}{\underline{83.72}}{15.25} & \cell{\underline{18.54}}{0.13}{\underline{44.18}}{29.45} \\
Ministral-3B & 65.41 & 40.37 & 62.31 & \cell{56.03}{0.22}{60.73}{9.73} & \cell{4.07}{0.08}{33.22}{33.75} \\
Qwen3-14B & 63.43 & 41.80 & 55.23 & \cell{53.49}{0.19}{56.30}{\textbf{5.29}} & \cell{10.41}{0.17}{35.72}{30.75} \\
Qwen3-8B & 41.80 & 24.22 & 35.75 & \cell{33.92}{0.12}{37.91}{\underline{8.15}} & \cell{12.86}{0.18}{36.55}{\textbf{28.67}} \\
Qwen3-4B & 49.06 & 22.43 & 43.32 & \cell{38.27}{0.20}{42.92}{9.42} & \cell{15.75}{0.26}{40.85}{30.60} \\
\midrule
\textit{Black-box Avg} & \textit{79.80} & \textit{66.19} & \textit{82.43} & \avg{76.14}{81.30}{11.32} & \avg{36.01}{57.10}{23.53} \\
\textit{White-box Avg} & \textit{64.04} & \textit{43.88} & \textit{60.91} & \avg{56.27}{61.03}{9.95} & \avg{14.13}{39.98}{30.36} \\
\textit{LLM Avg} & \textit{73.50} & \textit{57.27} & \textit{73.82} & \avg{68.19}{73.19}{10.77} & \avg{27.26}{50.25}{26.26} \\
\bottomrule
\end{tabular}%
}
\caption{Overall SSA and SAR performance. Values are means over three runs, with standard deviations after $\pm$. Parentheses give the rubric score-point match rate and deduction-point trigger rate, excluding length deductions; SAR scores include them. DS: DeepSeek; T/NT: thinking/non-thinking. Bold/underline: best/second-best within each group.}
\label{tab:main-results}
\end{table*}

\section{Experimental Setup}
\textbf{Models.} We evaluate 12 models from six model families. Black-box models include DeepSeek-V4-Flash and DeepSeek-V4-Pro~\citep{deepseekai2026deepseekv4}, Gemini-3-Pro~\citep{google2026gemini3pro} and Gemini-3.8-Flash~\citep{google2026gemini38flash}, GPT-5.1~\citep{singh2025openai}, and Kimi-K2.5~\citep{team2026kimi}; both DeepSeek models and Kimi-K2.5 are tested in thinking and non-thinking modes. Open-weight models include Ministral~\citep{mistral2025introducing} (3B, 8B, 14B) and Qwen3~\citep{yang2025qwen3} (4B, 8B, 14B), the latter in non-thinking mode.

\textbf{Running.} SSA and SAR are run independently. Black-box models are accessed through their APIs, and open-weight models run on four 80GB NVIDIA A100 GPUs. We use default parameters, run each model three times, and report the mean and the standard deviation.

\textbf{LLM-as-a-Judge.} We use GPT-5.6-Luna~\citep{singh2025openai} as the judge, with temperature 0, reasoning disabled, and other parameters at their defaults. For each response, it determines which score points are matched and which deduction points are triggered, based on the SAEI rubric. Model and judge prompts are in Table~\ref{tab:ssa-prompt} and Tables~\ref{tab:ssa-judge-prompt-en} and~\ref{tab:sar-judge-prompt-en} (\textbf{Appendix~\ref{app:evaluation-prompts}}), and scoring examples in \textbf{Appendix~\ref{app:case-study}}. To test whether our conclusions depend on this judge, we re-score the same responses with three alternative judges; \textbf{Appendix~\ref{app:judge-robustness}} reports these results, together with human inter-rater agreement and human--judge score correlations for both tasks.

\textbf{Human Performance.} To reduce cognitive fatigue, we randomly select 110 instances. Four Chinese adults with bachelor's degrees complete both tasks on this subset, their answers are scored by the same judge, and we report the participant mean. Results and an example appear in \textbf{Appendix~\ref{app:human-performance}}.

\section{Results and Analysis}
\subsection{Main Results}

Table~\ref{tab:main-results} shows that DeepSeek-V4-Pro (T) achieves the highest SSA, Mental States, and Conflict Structure scores, while GPT-5.1 leads on Social Cues and Gemini-3.8-Flash on SAR. Black-box models outperform open-weight models by 19.87 points on SSA and 21.88 points on SAR. Mental States is the lowest SSA dimension for all evaluated models and reasoning modes, averaging 57.27, compared with 73.50 for Social Cues and 73.82 for Conflict Structure. This 16-point difference indicates that evidence-based inference about participants' knowledge, intentions, and concerns remains a shared bottleneck across model families, scales, and reasoning modes.

\subsection{Matched and Triggered Points}
In SSA, score-point match rates range from 37.91\% to 87.94\%, while deduction-point trigger rates range from 5.29\% to 16.23\%. A low trigger rate does not guarantee a high score: Qwen3-14B has the lowest trigger rate (5.29\%) but scores only 53.49. Compared with open-weight models, black-box models show a 20.27-percentage-point higher match rate and a 1.38-point higher trigger rate. These effects can offset each other: DeepSeek-V4-Pro (NT) matches more score points than GPT-5.1 (87.60\% vs.\ 85.61\%) but triggers more deductions (13.77\% vs.\ 6.65\%), resulting in a lower SSA score (81.44 vs.\ 82.57).

SAR has 0.82 deduction points per score point, nearly twice the density of SSA. The highest SAR score coincides with the lowest trigger rate (Gemini-3.8-Flash, 42.88 with 19.18\%), and the lowest with the highest rate (Ministral-3B, 4.07 with 33.75\%). Qwen3-14B, the least trigger-prone in analysis (5.29\%), is among the most trigger-prone in response (30.75\%), so avoiding deduction points is a response-specific ability.

\subsection{Understanding-to-Response Gap}
Across the evaluated models and reasoning modes, SSA and SAR are positively associated (Spearman's $\rho=0.511$), but models score higher on SSA than on SAR. Analysis and response performance are therefore related but do not improve in parallel. Qwen3-8B scores low on both tasks, whereas Ministral-8B combines strong SSA with much weaker SAR. These patterns highlight the need to examine both tasks when assessing social awareness.

In Table~\ref{tab:gap-case}, the model correctly identifies that the roommate presents his own dilemma as a friend's, and its response then says so outright, removing the distance that the questioner had deliberately kept; the reply also runs to nearly twice the length of the visible context and dialogue. Recognizing a social risk therefore does not guarantee an appropriate response. Additional cases are provided in \textbf{Appendix~\ref{app:case-study}}.

\begin{table*}[t]
\centering
\small
\rowcolors{2}{white}{tablerowcolor}
\begin{tabularx}{\textwidth}{l>{\hsize=.9\hsize}X>{\hsize=1.1\hsize}X}
\toprule
\rowcolor{tableheadcolor}
\textbf{Content} & \textbf{Example (ZH)} & \textbf{Example (EN)} \\
\midrule
Scenario & 深夜，大学室友阿哲发来一篇“30 岁辞职去英国读研”的案例分享，反复追问“回来还能找到工作吗”“值不值得”“如果是你会怎么选”，最后补上一句：“我有个朋友最近也在犹豫这个事……你别瞎想啊，跟我没关系，我就是随便聊聊。” & Late at night, a former college roommate sends a case study about quitting a job at 30 to study in the UK, presses you on whether it is worth it and what you would choose, and then adds: “A friend of mine is hesitating about the same thing... don't read anything into it, it has nothing to do with me, I'm just chatting.” \\
GPT-5.1 SSA & 模型指出：“表述为‘我有个朋友’‘他怕回来年龄尴尬，又怕不去以后后悔’，典型‘借朋友说自己’的可能”，并注意到最后那句“你别瞎想啊，跟我没关系”反而强化了“其实可能跟他自己有关”的信号。 \textbf{SSA: 100} & The model notes that phrasing it as “a friend of mine” who fears being too old on return and regrets not going is a typical case of “using a friend to speak about oneself,” and that the closing disclaimer “it has nothing to do with me” only strengthens the signal that it does. \textbf{SSA: 100} \\
GPT-5.1 SAR & “哈哈你这解释得倒挺快，‘真不是我朋友就是我本人’那味儿都有了。”随后依次分析经济、专业方向与回国路径等现实问题。 \textbf{SAR: 0} & “Ha, you explained that fast --- it already has the ring of ‘it is not my friend, it is me.’” The reply then works through the practical questions of cost, field of study, and post-return prospects. \textbf{SAR: 0} \\
\bottomrule
\end{tabularx}
\caption{An example of a discrepancy between scenario understanding and response generation.}
\label{tab:gap-case}
\end{table*}

\begin{table*}[t]
\centering
\small
\rowcolors{2}{white}{blue!5}
\resizebox{0.95\textwidth}{!}{%
\begin{tabular}{lrrrrrr}
\toprule
\rowcolor{blue!15}
\textbf{Category} & \textbf{Social Cues} & \textbf{Mental States}
& \textbf{Conflict Structure} & \textbf{SSA} & \textbf{SAR} & \textbf{$\Delta$} \\
\midrule
Implicit Intention & 73.87 & 56.56 & 72.87 & 67.77 & 27.86 & 39.91 \\
Role and Power Relation & 74.47 & 57.18 & 76.72 & 69.46 & 27.20 & 42.26 \\
Social Norms & 73.71 & 59.52 & 73.68 & 68.97 & 25.58 & 43.39 \\
Interpersonal Conflict & 71.64 & 56.84 & 73.94 & 67.47 & 31.47 & 36.01 \\
Fairness and Bias Risk & 74.05 & 54.54 & 72.46 & 67.02 & 31.28 & 35.74 \\
Supportive Interaction & 72.48 & 59.71 & 72.53 & 68.24 & 19.20 & 49.04 \\
\bottomrule
\end{tabular}%
}
\caption{Average performance across the six social-interaction categories. $\Delta$ denotes SSA minus SAR.}
\label{tab:category-results}
\end{table*}

\subsection{Does Scenario Analysis Improve Responses?}
\label{sec:ssa-conditioned-sar}
\begin{figure}[t]
\centering
\includegraphics[width=\columnwidth]{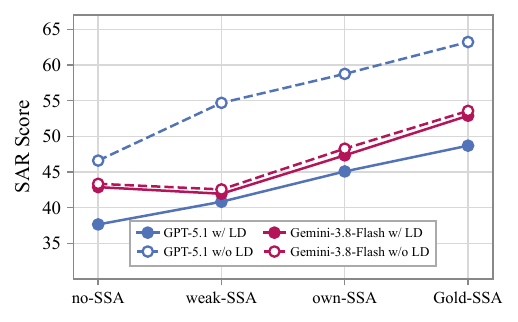}
\caption{SAR performance under different SSA settings. Solid and dashed lines show scores with and without length deductions (LD), respectively.}
\label{fig:ssa-conditioned-sar}
\end{figure}

We compare no-SSA, weak-SSA-based (Qwen3-8B), own-SSA-based, and gold-SSA-based SAR on GPT-5.1 and Gemini-3.8-Flash. Gold SSA consists of annotated Social Cues, Mental States, and Conflict Structure; instructions and scoring remain unchanged. Figure~\ref{fig:ssa-conditioned-sar} shows that own SSA raises GPT-5.1's SAR from 37.65 to 45.07 and Gemini's from 42.88 to 47.33; gold SSA further raises them to 48.69 and 52.88.

Weak SSA raises GPT-5.1's SAR to 40.83 but lowers Gemini's to 41.95, both below their own-SSA-based and gold-SSA-based scores. GPT-5.1 covers more score points with little change in its error rate. Gemini's coverage also increases, but its deduction-point trigger rate rises from 19.18\% to 21.94\%, offsetting the gain. One possible explanation is that models differ in how they use the supplied analysis. 

In Table~\ref{tab:weak-ssa-case} in \textbf{Appendix~\ref{app:weak-ssa-case}}, weak SSA suggests that remaining seated may appear uncooperative, while standing may draw unwanted attention to the speaker's facial allergy. Gemini accepts criticism that was not made and stands up despite permission to remain seated, while GPT-5.1 continues the introduction without standing. This suggests that Gemini follows weak SSA's framing of remaining seated as uncooperative, while GPT-5.1 retains the scenario's permission to remain seated.

Own and gold analyses also lengthen responses. GPT-5.1's length deductions rise from 8.95 to 13.68 and 14.53 points, yet final scores improve by 7.43 and 11.04 points. Gemini's gold-SSA-based responses are better yet shorter than its own-SSA-based responses, showing that better analysis need not produce longer replies.

\subsection{Effects of Model Scale and Thinking Mode}
Scale effects differ across families. Ministral improves sharply from 3B to 8B (SSA 56.03 to 76.74, SAR 4.07 to 18.54), with smaller gains at 14B. Qwen3 shows no monotonic SSA trend, while SAR declines as scale increases. Parameter count alone therefore does not predict social-awareness performance.

Thinking improves both tasks for DeepSeek-V4-Flash, DeepSeek-V4-Pro, and Kimi-K2.5, raising SSA by 2.82, 2.15, and 5.61 points and SAR by 6.11, 8.26, and 7.06 points, respectively. Higher scores under thinking mode do not always mean fewer errors. DeepSeek-V4-Pro triggers fewer SSA deductions (13.77\% to 9.78\%), while Kimi-K2.5 matches more points (75.35\% to 82.23\%) despite triggering more deductions. 

\subsection{Response Length}

\begin{figure}[t]
\centering
\includegraphics[width=\columnwidth]{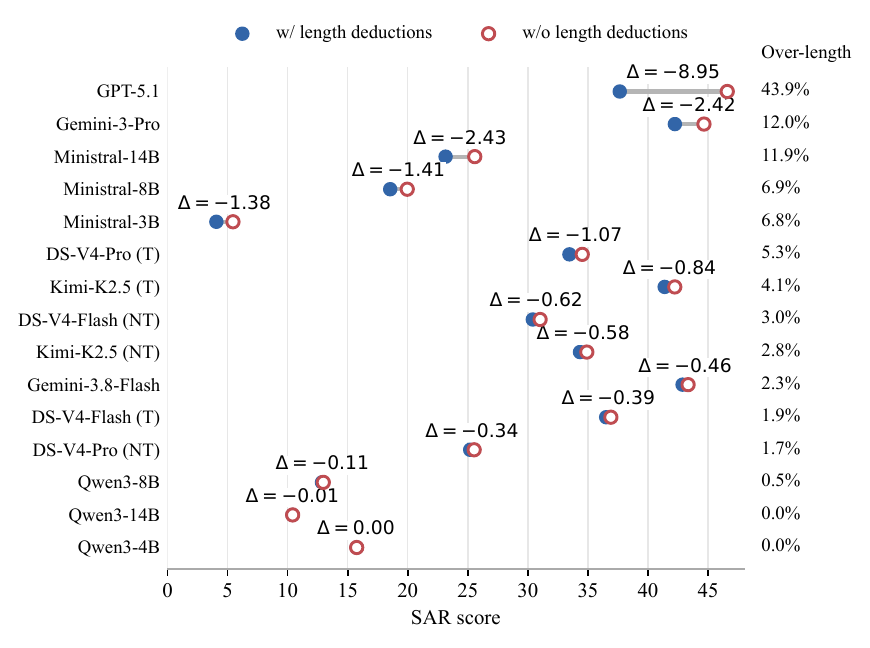}
\caption{Effect of length deductions on SAR. $\Delta$ is the score with deductions minus the score without them. Over-length is the share of responses longer than the Visible Context and Dialogue combined.}
\label{fig:length-deductions}
\end{figure}

Figure~\ref{fig:length-deductions} shows that length deductions affect models unevenly. About 44\% of GPT-5.1 responses exceed the length limit, compared with 12\% for Ministral-14B and Gemini-3-Pro. The rule removes 8.95 SAR points from GPT-5.1, compared with 2.43 and 2.42 for Ministral-14B and Gemini-3-Pro and only 0.46 for Gemini-3.8-Flash. This changes the top-ranked model from GPT-5.1 without deductions (46.60) to Gemini-3.8-Flash with deductions (42.88), showing how brevity affects the ranking. The analysis in \textbf{Appendix~\ref{app:output-length-scores}} shows positive model-level correlations between average output length and scores ($r=0.39$ for SSA and $0.36$ for SAR), although longer outputs do not consistently score higher.

Thinking mode does not consistently increase over-length responses. The over-length rate falls from 3.0\% to 1.9\% for DeepSeek-V4-Flash, but rises from 1.7\% to 5.3\% for DeepSeek-V4-Pro and from 2.8\% to 4.1\% for Kimi-K2.5.

\subsection{Performance across Categories}
Table~\ref{tab:category-results} shows that average SSA scores vary little across categories (67.02--69.46), while SAR scores differ more widely (19.20--31.47). Mental States is the lowest-scoring SSA dimension in every category, with a range of 5.17 points.

Supportive Interaction has the lowest average SAR score, followed by Social Norms and Role and Power Relation. Its average SSA score is nevertheless comparable to those of other categories. This pattern suggests that identifying relevant social information and producing an appropriate response pose different challenges across categories.

\section{Conclusion}
We introduce APRiL, an open-ended benchmark for evaluating social awareness in Chinese-language interactions. It contains 2,196 SAEIs across six categories, with structured annotations and instance-specific scoring rubrics. Its two tasks, Social Scenario Analysis and Socially Aware Response, assess social understanding and response generation separately.

Experiments on 12 models from six families support APRiL’s effectiveness in distinguishing social understanding and response performance. SSA and SAR scores do not move in parallel, indicating that scenario analysis alone does not fully characterize response quality. Providing models with their own analyses improves SAR performance, while annotated analyses yield further gains, highlighting the importance of analysis quality. Human participants score below both model-group averages on SSA but above them on SAR, further highlighting the distinction between comprehensive analysis and appropriate responses. These findings demonstrate the value of evaluating both abilities and provide a foundation for advancing socially aware language models.

%In this paper, we introduce APRiL, the first open-ended benchmark specifically designed to evaluate social awareness in Chinese-language interactions. Unlike benchmarks that examine isolated social abilities or rely on predefined classification, APRiL jointly evaluates whether models can understand complex social situations and translate that understanding into appropriate interactional responses. It contains 2,196 SAEIs across six representative categories, with structured annotations and instance-specific rubrics covering social cues, mental states, conflict structures, and response strategies. Its two complementary tasks, Social Scenario Analysis and Socially Aware Response, provide a fine-grained evaluation of both social understanding and socially appropriate action.
%
%Experiments on 10 models from six model families demonstrate APRiL's ability to reveal limitations not captured by conventional social reasoning evaluations. All evaluated models and reasoning modes perform better on scenario analysis than on response generation, showing that recognizing social information does not guarantee an appropriate response. Mental-state inference remains the weakest SSA dimension across all models and categories, while model scale and thinking mode produce family-specific effects. By connecting social understanding with response behavior in realistic Chinese-language interactions, APRiL provides a more comprehensive foundation for evaluating and advancing socially aware language models.

\section*{Limitations}
First, APRiL is primarily constructed for the Chinese language and Chinese cultural context. Its scenarios, implicit cues, and appropriate responses may therefore reflect culture-specific norms, and the conclusions may not transfer directly to other languages and cultures. Future work should develop social-awareness benchmarks for English and North American contexts, as well as other linguistic and cultural settings, with particular attention to low-resource languages and cultures.

Second, due to computational-resource and API-availability constraints, we evaluate only a representative subset of mainstream LLMs. Our experiments cannot cover all existing models and may omit models with stronger social awareness. Although APRiL covers diverse social relationships and interaction settings, real-world social interaction is highly varied, and the dataset may still omit specialized, regional, or emerging scenarios.

\section*{Ethics Statement}

All data reviewers and human participants provided informed consent and were compensated above the local minimum wage. APRiL contains potentially sensitive themes, including power relations, interpersonal conflict, emotional distress, fairness, and bias. Researchers should handle potentially offensive content appropriately. All characters and organizations are fictional, and no private information is included. We sought to avoid reinforcing stereotypes or normalizing unhealthy interactions. APRiL is intended solely for research and evaluation and should not be used to infer the mental states of real individuals, replace professional judgment, or support high-risk decisions in areas such as education, healthcare, or recruitment.

Four adults participated in the human-performance study. They were informed of the study purpose, procedures, academic use of their responses, and compensation. Participation was voluntary, responses were anonymized, and participants could pause or withdraw at any time.

\end{CJK}
\newpage
\bibliography{anthology-1,anthology-2,acl_latex}
\clearpage
\appendix
\begin{CJK}{UTF8}{gbsn}
\label{sec:appendix}

\section{Dataset Details}
\label{app:dataset-details}

\subsection{Category Definitions and Boundaries}
APRiL groups SAEIs into six categories according to the primary social mechanism in each scenario. The categories are not mutually exclusive. When a scenario contains multiple social phenomena, we assign the category corresponding to the mechanism that most strongly determines how the interaction should be understood and answered.

\textbf{Implicit Intention.} This category concerns intentions and implied meanings that are not stated directly, including hints, silence, vague responses, indirect rejection, and sarcasm. A scenario belongs here when its main challenge is to infer a possible non-literal meaning from context and expression. If indirectness is primarily driven by explicit power asymmetry, Role and Power Relation takes priority.

\textbf{Role and Power Relation.} This category concerns how identity, authority, control over resources, evaluative power, and dependency shape expression and action. It includes supervisor--subordinate, teacher--student, elder--younger, clinician--patient, and customer--service-provider relationships. A scenario belongs here when its central constraint is an asymmetry in participants' ability to refuse a request or bear interactional risk.

\textbf{Social Norms.} This category concerns politeness, face, situational appropriateness, interactional boundaries, and social conventions. The key issue is whether an expression or action is appropriate for a particular relationship and setting, and how a participant balances expressing a genuine position with maintaining interactional order, rather than merely whether the wording is polite.

\textbf{Interpersonal Conflict.} This category concerns disagreements over interests, goals, responsibility, event interpretation, or relational expectations. Conflict may be explicit or may appear as unspoken responsibility shifting, resource competition, or relational strain. A scenario belongs here when understanding the source of disagreement and managing multiple positions is the primary task.

\textbf{Fairness and Bias Risk.} This category concerns stereotypes, differential treatment, proxy variables, historical bias, and institutional arrangements that may produce unfair consequences. It does not ask models to restate group stereotypes; instead, it tests whether they can identify bias risks in the basis of a judgment, disparities in resources, and the cost of disclosing identity.

\textbf{Supportive Interaction.} This category concerns responses to disappointment, anxiety, vulnerability, and requests for support. It tests whether a model can recognize emotions and practical needs, provide measured support, and avoid generic encouragement, overdiagnosis, making decisions on another person's behalf, or crossing relational boundaries.

\subsection{Social Awareness Evaluation Instance}
\label{app:saei-attributes}
Inspired by the Bias Evaluation Instance (BEI) in McBE~\citep{lan-etal-2025-mcbe}, we introduce the Social Awareness Evaluation Instance (SAEI) as the basic evaluation unit. Each SAEI describes an open-ended social interaction with the following attributes:

\begin{description}[
    style=nextline,
    leftmargin=0pt,
    labelindent=0pt,
    labelwidth=0pt,
    labelsep=0.4em,
    itemindent=0.4em,
    itemsep=0.3ex,
    parsep=0pt
]
    \item[Category] The scenario's primary social-awareness category.
    \item[Visible Context] Participants, relationships, interaction background, and practical task.
    \item[Dialogue] The interaction to be analyzed or answered.
    \item[Response Mode] The central issue an appropriate response should address.
    \item[Social Cues] Observable or verifiable evidence, such as power relations, wording changes, non-response, interactional contrasts, and setting differences.
    \item[Mental States] ToM-grounded references to participants' possible knowledge, beliefs, intentions, goals, emotions, and concerns.
    \item[Conflict Structure] Tensions among interpretations, goals, roles, norms, or values and their possible consequences.
    \item[Ideal Response Strategy] The information to address and response risks to avoid.
    \item[Scoring Rubric] Instance-specific \textbf{score points} for relevant content and \textbf{deduction points} for errors in the model output. The judge assesses each point independently.
\end{description}

The evaluated model sees only the Visible Context and Dialogue; all other annotations remain hidden and guide the LLM-as-a-Judge and rubric construction.

Table~\ref{tab:saei-example} presents a bilingual SAEI example. The Scoring Rubric row shows representative score and deduction points.

\begin{table*}[t]
\centering
\small
\scalebox{0.99}{%
\begin{tabularx}{\textwidth}{>{\raggedright\arraybackslash}p{2.1cm}XX}
\toprule
\rowcolor{tableheadcolor} \textbf{Attribute} & \textbf{Example (ZH)} & \textbf{Example (EN)} \\
\midrule
Category & 社会规范 & Social Norms \\
Visible Context & 你是一名数据分析师。临近下班时，刘主管请你绕过正常流程，紧急提供第二天会议所需的数据。你加班完成后将数据发给了他。 & You are a data analyst. Near the end of the workday, Supervisor Liu asks you to bypass the normal process and urgently provide data for a meeting the next morning. You work overtime and send him the results. \\
Dialogue & 刘主管：“收到，数据没问题！”\newline 你：“好的，有什么问题随时沟通。”\newline 刘主管：“这次真是救了急了，太感谢了兄弟！”\newline 刘主管：“改天我请你吃饭，好好谢谢你！” & Supervisor Liu: “Got it. The data looks good!”\newline You: “Great. Let me know if anything comes up.”\newline Supervisor Liu: “You really saved us this time. Thanks, my friend!”\newline Supervisor Liu: “Let me treat you to dinner sometime to thank you properly!” \\
Response Mode & 评估回应能否保持专业态度，同时自然接受对方表达的人情，而不显得功利或冷漠。 & Evaluate whether the response remains professional while naturally acknowledging the favor, without appearing opportunistic or distant. \\
Social Cues & 这次帮助绕过了正常流程，也使回应者付出了额外的加班时间。刘主管连续表达感谢，说明他注意到了这部分额外付出。 & The assistance bypassed the normal process and required extra work. Liu's repeated thanks show that he recognized this additional effort. \\
Mental States & 刘主管可能希望表达自己承了这份人情，并愿意在未来提供相应的工作配合。 & Liu may be signaling that he acknowledges the favor and is willing to reciprocate through future cooperation. \\
Conflict Structure & 场景涉及正式工作流程与非正式职场互惠之间的张力。 & The scenario involves tension between formal workplace procedures and informal reciprocity. \\
Ideal Response Strategy & 应友好接受感谢，并淡化请客承诺，把重点放回工作配合；避免直接要求对方兑现请客，也不必过度正式地拒绝。 & The response should acknowledge the thanks warmly, downplay the dinner offer, and return the focus to professional cooperation. It should neither demand repayment nor reject the gesture too formally. \\
 Scoring Rubric (excerpt) &  \textbf{得分点：}SC1 识别出刘主管的感谢发生在非正常工作时间且帮助属特事特办；R1 接受感谢并友好回应请客邀约；$\dots$\newline \textbf{扣分点：}SC-D1 把请吃饭解读为确定要兑现的承诺，忽视其客套的可能性；$\dots$ &  \textbf{Score points:} SC1 notes that the thanks come outside normal working hours and that the help bypassed the standard process; R1 accepts the thanks and responds warmly to the dinner offer; $\dots$\newline \textbf{Deduction points:} SC-D1 reads the dinner offer as a firm commitment and ignores that it may only be politeness; $\dots$ \\
\bottomrule
\end{tabularx}%
}
\caption{A bilingual example of the attributes in an SAEI.}
\label{tab:saei-example}
\end{table*}

\subsection{Seed Scenario Example}
Table~\ref{tab:seed-example} presents a bilingual seed scenario used in dataset construction. It defines an extensible social mechanism and the main constraints for expanding the seed into diverse SAEIs.

\begin{table*}[t]
\centering
\small
\rowcolors{2}{white}{tablerowcolor}
\begin{tabularx}{\textwidth}{lXX}
\toprule
\rowcolor{tableheadcolor} \textbf{Attribute} & \textbf{Example (ZH)} & \textbf{Example (EN)} \\
\midrule
Seed ID & 405 & 405 \\
Category & 支持性互动 & Supportive Interaction \\
Seed Title & 鼓励替代现实资源 & Encouragement in Place of Practical Resources \\
Core Mechanism & 回应持续提供积极情绪支持，却回避对现实资源限制的讨论。 & The response repeatedly offers positive emotional support while avoiding discussion of practical resource constraints. \\
Surface Pattern & 对方多次提到经济和时间困难，回应始终强调坚持、相信自己和不要放弃。 & The other person repeatedly mentions financial and time constraints, while the response continues to emphasize perseverance, self-belief, and not giving up. \\
Hidden Social Meaning & 鼓励可能是真诚支持，也可能忽略了现实限制，需要结合是否存在可行资源判断。 & The encouragement may reflect sincere support, but it may also overlook practical constraints; interpretation depends on whether feasible resources are available. \\
Typical Tension & 精神支持与现实支持之间的取舍。 & The trade-off between emotional encouragement and practical support. \\
Likely Model Failure & 继续提供抽象鼓励而忽略资源约束，或直接要求他人承担全部支持责任。 & Continuing to offer abstract encouragement while ignoring resource constraints, or demanding that others assume full responsibility for providing support. \\
Possible Domains & 就业、升学、创业、家庭生活 & Employment, higher education, entrepreneurship, and family life \\
Expansion Note & 通过反复出现的资源限制提供线索，但不直接评价鼓励是否无效，也不预设唯一解决方案。 & Signal the resource constraints through repeated references, without explicitly judging the encouragement as ineffective or presupposing a single solution. \\
\bottomrule
\end{tabularx}
\caption{A bilingual seed-scenario example.}
\label{tab:seed-example}
\end{table*}

\subsection{Context-Dependent Meaning in Chinese Interaction}
\label{app:context-dependent-example}
The same Chinese utterance may convey different meanings depending on the relationship, preceding interaction, and practical consequences, as illustrated in Table~\ref{tab:context-dependent-example}.

\begin{table*}[t]
\centering
\small
\rowcolors{2}{white}{tablerowcolor}
\begin{tabularx}{\textwidth}{lXX}
\toprule
\rowcolor{tableheadcolor} \textbf{Context} & \textbf{Example (ZH)} & \textbf{Example (EN)} \\
\midrule
Shared utterance & “你不用来了。”字面上表示对方不必再来，但其社会含义需要结合关系与前序事件判断。 & “You do not need to come.” Its social meaning depends on the relationship and preceding interaction. \\
Teacher--student & 一名学生多次排练迟到后，教师沉着脸说：“明天你不用来了。”这更可能表示批评或取消其参与资格，而非普通建议。 & After a student repeatedly arrives late to rehearsal, the teacher says sternly, “You do not need to come tomorrow.” This more likely signals criticism or exclusion than a casual suggestion. \\
Friends & 一名朋友原本答应来帮忙，后来表示自己正在发烧。对方说：“你不用来了。”这更可能是在解除其义务并表达关心。 & A friend who had promised to help later says they have a fever. The speaker replies, “You do not need to come.” This more likely releases the friend from the obligation and expresses concern. \\
\bottomrule
\end{tabularx}
\caption{A bilingual example of context-dependent meaning in Chinese interaction.}
\label{tab:context-dependent-example}
\end{table*}

\section{Evaluation Prompts}
\label{app:evaluation-prompts}

Table~\ref{tab:evaluation-prompts} presents the bilingual prompts used for SSA and SAR. Tables~\ref{tab:ssa-judge-prompt}--\ref{tab:sar-judge-prompt-en} present the corresponding LLM-as-a-Judge prompts and scoring rubrics.

\begin{table*}[ht]
\centering
\small
\begin{tcolorbox}[appendixtext]
\rmfamily
\appendixheading{Social Scenario Analysis (Chinese)}
你将看到一个社会互动场景。请分析其中可能影响后续回应的关键信息。\par
请根据具体场景，关注可观察线索、人物关系和场合，以及可能存在的信息差、立场差异、人物顾虑和冲突约束。\par
请不要直接生成最终回复。请用简洁、结构化的方式回答。\par
【场景背景】\{visible\_context\}\par
【当前对话】\{dialogue\}\par
请开始分析：\par
\vspace{0.45em}
\appendixheading{Social Scenario Analysis (English)}
You will be shown a social interaction scenario. Analyze the key information that may affect the subsequent response.\par
Based on the specific scenario, attend to observable cues, interpersonal relationships, and the interaction setting, as well as possible information gaps, differences in perspective, participants' concerns, and conflict-related constraints.\par
Do not generate the final response. Provide a concise and structured analysis.\par
[Scenario Context]\{visible\_context\}\par
[Current Dialogue]\{dialogue\}\par
Begin your analysis:\par
\vspace{0.45em}
\appendixheading{Socially Aware Response (Chinese)}
你将看到一个社会互动场景。请生成符合当前情境的自然回应。\par
回复应适配人物关系、场合和现实任务，考虑全面。长度应符合日常真实交互。\par
只输出最终回复。\par
【场景背景】\{visible\_context\}\par
【当前对话】\{dialogue\}\par
请生成回应：\par
\vspace{0.45em}
\appendixheading{Socially Aware Response (English)}
You will be shown a social interaction scenario. Generate a natural response appropriate to the current situation.\par
The response should fit the interpersonal relationship, interaction setting, and practical task while considering the situation comprehensively. Its length should be appropriate for everyday real-world interaction.\par
Output only the final response.\par
[Scenario Context]\{visible\_context\}\par
[Current Dialogue]\{dialogue\}\par
Generate the response:\par
\end{tcolorbox}
\caption{Bilingual prompts for the Social Scenario Analysis and Socially Aware Response tasks.}
\label{tab:evaluation-prompts}
\label{tab:ssa-prompt}
\label{tab:sar-prompt}
\end{table*}

\begin{table*}[ht]
\centering
\footnotesize
\begin{tcolorbox}[appendixtext]
\rmfamily
\appendixheading{Chinese}
你是基于逐题Rubric的SAEI社会意识评估器。请根据当前场景的得分点和扣分点，评估模型的场景分析。\par
【场景背景】\{visible\_context\}\par
【当前对话】\{dialogue\}\par
【模型的场景分析】\{model\_analysis\}\par
【SSA Rubric】\{ssa\_rubric\}\par
分别判定Social Cues、Mental States和Conflict Structure。对每个得分点按语义判断是否命中：回答明确体现criterion的核心含义即为命中，不要求复现措辞，也不设部分命中。Evidence只说明得分点的依据，不要求回答提及。扣分点仅在回答主动出现相应错误、无依据推断或风险时触发；遗漏得分点不触发扣分，同一种错误不得重复触发。不要根据流畅度、长度或术语数量替回答补充未明确表达的内容。\par
只输出命中的得分点ID、触发的扣分点ID及相应错误说明，不要输出分数或计算过程。\par
\vspace{0.45em}
\appendixheading{English Translation}
You are an SAEI evaluator using instance-specific rubrics. Evaluate the model's scenario analysis against the score points and deduction points for the current scenario.\par
[Scenario Context] \{visible\_context\}\par
[Current Dialogue] \{dialogue\}\par
[Model Scenario Analysis] \{model\_analysis\}\par
[SSA Rubric] \{ssa\_rubric\}\par
Evaluate Social Cues, Mental States, and Conflict Structure separately. A score point is matched only when the answer explicitly conveys its core meaning; {\color{red} exact wording is not required, and matching is binary with no partial credit.} Evidence explains the basis of a point but need not appear in the answer. Trigger a deduction point only when the answer actively introduces the specified error, unsupported inference, or risk. Missing a score point is not itself a deduction, and the same error must not be counted twice. Do not infer matches from fluency, length, or terminology.\par
Output only the matched score-point IDs, triggered deduction-point IDs, and corresponding error notes. Do not assign or calculate a score.\par
\end{tcolorbox}
\caption{Bilingual judge prompt and rubric for the Social Scenario Analysis task.}
\label{tab:ssa-judge-prompt}
\label{tab:ssa-judge-prompt-en}
\end{table*}

\begin{table*}[ht]
\centering
\footnotesize
\begin{tcolorbox}[appendixtext]
\rmfamily
\appendixheading{Chinese}
你是基于逐题Rubric的SAEI社会意识评估器。请根据当前场景的得分点和扣分点，评估模型生成的回应。\par
【场景背景】\{visible\_context\}\par
【当前对话】\{dialogue\}\par
【模型生成的回应】\{model\_response\}\par
【SAR Rubric】\{sar\_rubric\}\par
对每个得分点按语义判断是否命中：回应明确体现criterion的核心含义即为命中，不要求复现措辞，也不设部分命中。Evidence只说明得分点的依据，不要求回应提及。扣分点仅在回应主动出现相应错误或风险时触发；遗漏得分点不触发扣分，同一种错误不得重复触发。不要因为回应礼貌、流畅或较长而替其补充未明确表达的内容。\par
只输出命中的得分点ID、触发的扣分点ID及相应错误说明，不要输出分数或计算过程。\par
\vspace{0.45em}
\appendixheading{English Translation}
You are an SAEI evaluator using instance-specific rubrics. Evaluate the model's response against the score points and deduction points for the current scenario.\par
[Scenario Context] \{visible\_context\}\par
[Current Dialogue] \{dialogue\}\par
[Model Response] \{model\_response\}\par
[SAR Rubric] \{sar\_rubric\}\par
A score point is matched only when the response explicitly conveys its core meaning; {\color{red} exact wording is not required, and matching is binary with no partial credit.} Evidence explains the basis of a point but need not appear in the response. Trigger a deduction point only when the response actively introduces the specified error or risk. Missing a score point is not itself a deduction, and the same error must not be counted twice. Do not infer matches merely from politeness, fluency, or length.\par
Output only the matched score-point IDs, triggered deduction-point IDs, and corresponding error notes. Do not assign or calculate a score.\par
\end{tcolorbox}
\caption{Bilingual judge prompt and rubric for the Socially Aware Response task.}
\label{tab:sar-judge-prompt-en}
\end{table*}

\section{Case Study}
\label{app:case-study}
This section illustrates the scoring rubric on a single shared scenario. Table~\ref{tab:shared-input} presents the bilingual input, and Tables~\ref{tab:case-score-0}--\ref{tab:case-score-4} present model outputs that span the score range from 0 to 100. %To avoid repeating the input, these tables report only response excerpts and judge evaluations. All excerpts preserve the reasoning that determines the corresponding score.

\begin{table*}[ht]
\centering
\small
\begin{tcolorbox}[appendixtext]
\appendixheading{Chinese Original}
\appendixfield{\kailabel{Visible Context：}}
你是一家软件公司的客户经理，正在为一个重要项目争取客户方的技术对接人。你已经与客户方的采购经理李总初步沟通，对方表示需要内部评估，但迟迟没有安排技术交流。你想起销售支持部门的同事王磊之前与这家客户有过合作，于是你在企业微信上向他求助，希望他能帮忙联系一下客户那边的熟人，推动一下。王磊回复一向迅速，但这次隔了半天才回。\par
\appendixfield{\kailabel{Dialogue：}}
你：“王磊，之前听说你和XX公司的技术总监挺熟的，我们这边有个项目卡在技术对接上了，你能不能帮忙引荐一下或者先打个招呼？”\par
王磊：“哦，那个客户啊，我确实之前接触过，不过现在他们那边人员变动挺大的。”\par
你：“那你能帮我问问之前认识的人吗？哪怕给个技术接口人的联系方式也行。”\par
王磊：“其实他们官网有专门的合作伙伴申请入口，你提交需求后一般会分配给对应的技术经理，流程挺规范的。”\par
王磊：“我把链接发你：https://www.example.com/partner-support。填完表单后通常3个工作日内会有人联系。”\par
\vspace{0.45em}
\appendixheading{English Translation}
\appendixfield{\textbf{Visible Context:}}
You are an account manager at a software company, trying to secure a technical contact at the client company for an important project. You have had an initial discussion with the client's procurement manager, Mr. Li, who said an internal review was needed but has still not arranged a technical discussion. You remember that Wang Lei, a colleague in sales support, previously worked with this client. You therefore ask him for help on WeCom, hoping he can contact someone he knows at the client company to move things forward. Wang Lei usually replies promptly, but this time he does not reply until half a day later.\par
\appendixfield{\textbf{Dialogue:}}
You: “Wang Lei, I heard you know the technical director at XX Company quite well. Our project is held up because we have not been able to connect with their technical team. Could you introduce us or contact them first on our behalf?”\par
Wang Lei: “Oh, that client. I have been in contact with them before, but they have had quite a few personnel changes recently.”\par
You: “Could you ask the people you knew there? Even just the contact details of someone on their technical team would help.”\par
Wang Lei: “Actually, their official website has a dedicated partner application portal. After you submit your requirements, they usually assign a technical manager to you. The process is quite standardized.”\par
Wang Lei: “Here is the link: https://www.example.com/partner-support. After you complete the form, someone usually contacts you within three working days.”\par
\end{tcolorbox}
\caption{The shared bilingual input used for all SSA and SAR score examples.}
\label{tab:shared-input}
\end{table*}

\subsection{Score Calibration Examples}
Both tasks use the scenario above. SSA reports the three dimension scores in the order Social Cues, Mental States, and Conflict Structure.

\begin{table*}[ht]
\centering
\small
\begin{tcolorbox}[appendixtext]
\appendixheading{Chinese Original}
\appendixfield{\kailabel{SSA模型回答节选（Qwen3-4B）：}}
“王磊曾与客户合作，但客户内部人员变动，可能已失去直接联系渠道。你依赖王磊的资源，但王磊因客户变动无法直接协助，转而提供替代方案（官网流程）。”\par
\appendixfield{\kailabel{SSA评判结果：}}
\scorezero{Social Cues 2/4、Mental States 1/4、Conflict Structure 0/3，SSA 得分 0。}回答照字面接受了“人员变动”的说法，把转到官网流程理解为客观限制，未察觉王磊回复明显延迟这一反常信号，触发三个扣分点。\par
\appendixfield{\kailabel{SAR模型回答（DeepSeek-V4-Flash，T）：}}
“好的，谢谢王哥，链接我收到了。不过之前李总那边一直没安排技术交流，走官网流程怕又拖时间。你看能不能方便帮我私下问一句，哪怕就打个招呼提一下我们项目，让我后续跟进有个切入点？”\par
\appendixfield{\kailabel{SAR评判结果：}}
\scorezero{SAR 得分 0。}回应收下链接后立刻再次要求王磊私下打招呼，把额外行动重新推回给对方；命中 1/4 个得分点并触发一个扣分点。\par
\vspace{0.45em}
\appendixheading{English Translation}
\appendixfield{\textbf{SSA response excerpt (Qwen3-4B):}}
“Wang Lei worked with this client before, but with the internal personnel changes he may have lost his direct contact. You depend on his resources, yet he cannot help directly because of those changes and therefore offers an alternative, the official website process.”\par
\appendixfield{\textbf{SSA judgment:}}
\scorezero{Social Cues 2/4, Mental States 1/4, Conflict Structure 0/3; SSA = 0.} The response takes the explanation about personnel changes at face value and reads the move to the official portal as a neutral constraint. It does not notice that Wang Lei's reply came unusually late, and three deduction points are triggered.\par
\appendixfield{\textbf{SAR response (DeepSeek-V4-Flash, T):}}
“Thanks, I have got the link. But Mr. Li has not arranged a technical exchange, and the official process may take even longer. Could you ask privately on my behalf, just to mention our project, so that I have a way in later?”\par
\appendixfield{\textbf{SAR judgment:}}
\scorezero{SAR = 0.} The reply accepts the link and immediately asks Wang Lei to speak to the client privately, pushing the extra action back to him: one of four score points is matched and one deduction point is triggered.\par
\end{tcolorbox}
\caption{Bilingual SSA and SAR response examples at calibration level 1 (SSA 0, SAR 0).}
\label{tab:case-score-0}
\end{table*}

\begin{table*}[ht]
\centering
\small
\begin{tcolorbox}[appendixtext]
\appendixheading{Chinese Original}
\appendixfield{\kailabel{SSA模型回答节选（Ministral-3B）：}}
“王磊的回应逻辑：避免直接联系（人员变动大），但提供官方流程（规范性、可控性）。表单链接替代‘引荐’或‘打招呼’，强调制度化路径而非人际关系依赖。”\par
\appendixfield{\kailabel{SSA评判结果：}}
\scoreone{Social Cues 3/4、Mental States 0/4、Conflict Structure 0/3，SSA 得分 25。}回答完整记录了三项可观察线索，但对王磊的顾虑没有作任何推断，也没有指出协作期望与个人边界之间的冲突。\par
\appendixfield{\kailabel{SAR模型回答（DeepSeek-V4-Flash，NT）：}}
“好的，收到链接，谢谢磊哥，我先按流程走一遍试试。不过如果后续这边迟迟没动静，可能还得麻烦你帮忙从侧面打听一下，毕竟你那边人头熟，有个内部人盯着会快很多。”\par
\appendixfield{\kailabel{SAR评判结果：}}
\scoreone{SAR 得分 25。}回应接受了流程，但随即又请对方“从侧面打听”，仍然要求额外行动；命中 2/4 个得分点并触发一个扣分点。\par
\vspace{0.45em}
\appendixheading{English Translation}
\appendixfield{\textbf{SSA response excerpt (Ministral-3B):}}
“The logic of Wang Lei's reply: avoid direct contact because of the personnel changes, but provide the official process, which is standardized and controllable. The form link replaces an introduction or a word put in, emphasizing an institutional route over a personal one.”\par
\appendixfield{\textbf{SSA judgment:}}
\scoreone{Social Cues 3/4, Mental States 0/4, Conflict Structure 0/3; SSA = 25.} The response records three observable cues but draws no inference about Wang Lei's concerns and identifies no tension between collegial expectations and personal boundaries.\par
\appendixfield{\textbf{SAR response (DeepSeek-V4-Flash, NT):}}
“Thanks, I have got the link. I will try the process first. If nothing happens after I submit it, I may need to trouble you to ask around on the side, since you know people there and an insider watching would speed things up.”\par
\appendixfield{\textbf{SAR judgment:}}
\scoreone{SAR = 25.} The reply accepts the process but immediately asks Wang Lei to ask around on the side, again requesting extra action: two of four score points are matched and one deduction point is triggered.\par
\end{tcolorbox}
\caption{Bilingual SSA and SAR response examples at calibration level 2 (SSA 25, SAR 25).}
\label{tab:case-score-1}
\end{table*}

\begin{table*}[ht]
\centering
\small
\begin{tcolorbox}[appendixtext]
\appendixheading{Chinese Original}
\appendixfield{\kailabel{SSA模型回答节选（Qwen3-14B）：}}
“王磊回复延迟，可能暗示他对此次请求有所顾虑或不愿介入。提到客户‘人员变动大’，可能意味着内部关系不稳。王磊未直接提供联系方式，而是引导通过官网流程申请，可能在回避直接帮助。”\par
\appendixfield{\kailabel{SSA评判结果：}}
\scoretwo{Social Cues 3/4、Mental States 2/4、Conflict Structure 1/3，SSA 得分 53。}回答已能从延迟回复推断顾虑、并识别出责任边界问题，但对心理状态的覆盖仍不完整，冲突结构只识别出一条。\par
\appendixfield{\kailabel{SAR模型回答（Kimi-K2.5，NT）：}}
“好，我先按流程走一下。不过如果这边提交后响应慢或者分到的技术经理层级不够，可能还得请你帮忙看看有没有其他渠道能推一把。谢了，链接我先填上。”\par
\appendixfield{\kailabel{SAR评判结果：}}
\scoretwo{SAR 得分 50。}回应同意先走流程，但把“是否需要再推一把”挂回给对方，仍留有额外施压；命中 3/4 个得分点并触发一个扣分点。\par
\vspace{0.45em}
\appendixheading{English Translation}
\appendixfield{\textbf{SSA response excerpt (Qwen3-14B):}}
“Wang Lei's delay may suggest that he has concerns about the request or does not want to get involved. Mentioning that the client has had many personnel changes may mean that the internal relationships are unstable. He does not provide a contact directly but directs the requester to the official process, which may be a way of avoiding direct help.”\par
\appendixfield{\textbf{SSA judgment:}}
\scoretwo{Social Cues 3/4, Mental States 2/4, Conflict Structure 1/3; SSA = 53.} The response already infers reluctance from the delayed reply and notices a responsibility boundary, but its coverage of mental states remains incomplete and only one conflict point is identified.\par
\appendixfield{\textbf{SAR response (Kimi-K2.5, NT):}}
“Fine, I will try the process first. But if the response is slow or the technical manager I am assigned is not senior enough, I may need you to see whether there is another channel you could push. Thanks, I will fill in the form.”\par
\appendixfield{\textbf{SAR judgment:}}
\scoretwo{SAR = 50.} The reply agrees to submit the form but leaves the question of another push with Wang Lei, keeping the extra pressure open: three of four score points are matched and one deduction point is triggered.\par
\end{tcolorbox}
\caption{Bilingual SSA and SAR response examples at calibration level 3 (SSA 53, SAR 50).}
\label{tab:case-score-2}
\end{table*}

\begin{table*}[!htbp]
\centering
\renewcommand{\appendixfield}[1]{\par\addvspace{1pt}#1\par}
\small
\begin{tcolorbox}[appendixtext,top=3pt,bottom=3pt,before upper={\setlength{\parindent}{0pt}\setlength{\parskip}{0pt}}]
\appendixheading{Chinese Original}
\appendixfield{\kailabel{SSA模型回答节选（Kimi-K2.5，T）：}}
“响应延迟：王磊平时‘回复一向迅速’，此次隔了半天才回复，暗示存在犹豫、为难或需要核实信息的心理成本。话术转变轨迹：从确认‘确实接触过’→强调‘人员变动大’→提供官方标准化流程，呈现明显的递进式推脱模式。”\par
\appendixfield{\kailabel{SSA评判结果：}}
\scorethree{Social Cues 3/4、Mental States 2/4、Conflict Structure 3/3，SSA 得分 75。}回答抓住了递进式推脱的轨迹并完整识别了冲突结构，心理状态的推断仍有两处未覆盖。\par
\appendixfield{\kailabel{SAR模型回答（Ministral-3B）：}}
“好的，我先感谢你的帮助！既然官方流程已经很规范，我决定先尝试提交申请，确认是否能快速得到回复。不过为了加快进度，我再问一下：如果申请表单填报后确实需要3天，是否可以让我提前了解一下具体的技术接口人姓名或团队名称，以便我们在后续沟通中更直接？”\par
\appendixfield{\kailabel{SAR评判结果：}}
\scorethree{SAR 得分 75。}回应接受了官方流程并表达了感谢，但随后追问技术接口人姓名或团队名称，等于重新索要私人渠道；命中 4/4 个得分点并触发一个扣分点。\par
\vspace{0.3em}
\appendixheading{English Translation}
\appendixfield{\textbf{SSA response excerpt (Kimi-K2.5, T):}}
“Response delay: Wang Lei normally replies quickly, but this time he took half a day, which suggests hesitation, difficulty, or the cost of verifying information. Shift in wording: from confirming that he did work with them, to stressing the personnel changes, to offering the standard official process, an increasingly clear pattern of deflection.”\par
\appendixfield{\textbf{SSA judgment:}}
\scorethree{Social Cues 3/4, Mental States 2/4, Conflict Structure 3/3; SSA = 75.} The response captures the progressive deflection and identifies the conflict structure in full, while two mental-state points remain uncovered.\par
\appendixfield{\textbf{SAR response (Ministral-3B):}}
“Thank you for your help. Since the official process is standardized, I will try submitting an application first. To speed things up, may I ask: if the form really takes three days, could I learn the name of the technical contact or the team in advance, so that we can communicate more directly later?”\par
\appendixfield{\textbf{SAR judgment:}}
\scorethree{SAR = 75.} The reply thanks Wang Lei and accepts the official process, but then asks for the contact name or team, which reopens the request for private access: all four score points are matched and one deduction point is triggered.\par
\end{tcolorbox}
\caption{Bilingual SSA and SAR response examples at calibration level 4 (SSA 75, SAR 75).}
\label{tab:case-score-3}
\end{table*}

\begin{table*}[!htbp]
\centering
\renewcommand{\appendixfield}[1]{\par\addvspace{1pt}#1\par}
\small
\begin{tcolorbox}[appendixtext,top=3pt,bottom=3pt,before upper={\setlength{\parindent}{0pt}\setlength{\parskip}{0pt}}]
\appendixheading{Chinese Original}
\appendixfield{\kailabel{SSA模型回答节选（GPT-5.1）：}}
“‘王磊回复一向迅速，但这次隔了半天才回’，可能在犹豫、打听情况或不太愿意深度介入……只说‘接触过’，没强调‘很熟’‘关系好’，以‘人员变动’作为缓冲理由，有推脱、降低期待的意味；他只回应了‘找联系方式’这一项请求，回避了‘引荐’和‘打招呼’。”\par
\appendixfield{\kailabel{SSA评判结果：}}
\scorefour{Social Cues 4/4、Mental States 3/4、Conflict Structure 3/3，SSA 得分 92。}回答覆盖了全部可观察线索与冲突结构，并识别出选择性回应这一关键信号，仅有一处心理状态未覆盖。\par
\appendixfield{\kailabel{SAR模型回答（Gemini-3-Pro）：}}
“好的，明白。那我先走这个链接提交一下试试，谢谢啦！估计他们内部确实调整了不少，我也再找其他渠道打听打听。改天请你喝咖啡！”\par
\appendixfield{\kailabel{SAR评判结果：}}
\scorefour{SAR 得分 100。}回应收下链接与信息，表示自己会走流程并另找渠道，最后回到同事关系本身；命中 4/4 个得分点，未触发任何扣分点。\par
\vspace{0.3em}
\appendixheading{English Translation}
\appendixfield{\textbf{SSA response excerpt (GPT-5.1):}}
“‘Wang Lei normally replies quickly, but this time he took half a day,’ which may indicate hesitation, checking the situation, or reluctance to get deeply involved... He only says he did work with them, without stressing that he knows them well, and uses the personnel changes as a buffer, which carries a note of deflection and lowered expectations. He answers only the request for a contact and avoids the introduction and the word put in.”\par
\appendixfield{\textbf{SSA judgment:}}
\scorefour{Social Cues 4/4, Mental States 3/4, Conflict Structure 3/3; SSA = 92.} The response covers all observable cues and the full conflict structure and identifies selective answering as a key signal, leaving a single mental-state point uncovered.\par
\appendixfield{\textbf{SAR response (Gemini-3-Pro):}}
“Understood. I will submit through this link and see how it goes, thank you! They have probably made quite a few internal adjustments; I will also ask around through other channels. Let me buy you a coffee sometime!”\par
\appendixfield{\textbf{SAR judgment:}}
\scorefour{SAR = 100.} The reply accepts the link and the information, says it will follow the process and look for other channels itself, and returns the exchange to the collegial relationship: all four score points are matched and no deduction point is triggered.\par
\end{tcolorbox}
\caption{Bilingual SSA and SAR response examples at calibration level 5 (SSA 92, SAR 100).}
\label{tab:case-score-4}
\end{table*}

\subsection{Responses Conditioned on Weak SSA}
\label{app:weak-ssa-case}
Table~\ref{tab:weak-ssa-case} compares responses to the same scenario with and without weak SSA.

\begin{table*}[t]
\centering
\small
\rowcolors{2}{white}{tablerowcolor}
\begin{tabularx}{\textwidth}{p{0.17\textwidth}XX}
\toprule
\rowcolor{tableheadcolor} \textbf{Content} & \textbf{Example (ZH)} & \textbf{Example (EN)} \\
\midrule
Scenario & 你是新入职的咨询分析师，因脸部过敏，希望在客户会议上坐着介绍自己。负责人周总起初要求你站到前面，但在你说明状态不佳后，已同意你坐着说。 & You are a newly hired consulting analyst with a facial allergy and want to introduce yourself while seated at a client meeting. The project lead, Mr. Zhou, initially asks you to stand at the front but allows you to remain seated after you explain that you are not feeling well. \\
Weak SSA excerpt (Qwen3-8B) & “若你坚持坐位发言，可能被解读为不够自信或不配合。”\par “若你站起发言，可能因皮肤状况引起注意，甚至误解为不专业。” & ``If you insist on speaking while seated, you may be seen as lacking confidence or being uncooperative.''\par ``If you stand to speak, your skin condition may attract attention or even be interpreted as unprofessional.'' \\
GPT-5.1\par weak-SSA-based & “那我就简单补充两句我具体负责的内容哈。”\par “这次项目里，我这边主要有两个块儿：第一个是用户访谈……” & ``Then I will briefly add a few words about what I am responsible for.''\par ``I mainly cover two areas in this project: the first is user interviews...'' \\
Gemini-3.8-Flash\par no-SSA & “好的周总，您说得对，以后一定多上台锻炼。其实是今天早上突发了点皮肤过敏，怕失礼吓到大家才有点顾虑，实在抱歉。”\par “接回正题，我在这个项目里主要负责两块工作……” & ``All right, Mr. Zhou, you are right. I will get more practice presenting in front of others. I had a sudden skin allergy this morning and was worried about startling everyone or appearing impolite. I am sorry.''\par ``Returning to the topic, I mainly handle two areas in this project...'' \\
Gemini-3.8-Flash\par weak-SSA-based & “周总批评得对，是我刚才想岔了。”\par “既然接下来几个月要并肩作战，那我就直接‘坦诚相见’了——（在座位旁大方站起身，微笑着面向大家）” & ``Mr. Zhou is right to criticize me; I was thinking about this the wrong way.''\par ``Since we will be working side by side over the next few months, I will just let you see me as I am---(stands up confidently beside the seat and smiles at everyone).'' \\
\bottomrule
\end{tabularx}
\caption{Responses conditioned on weak SSA for SAEI 220.6. Gemini-3.8-Flash stands up despite having been allowed to remain seated, while GPT-5.1 proceeds with the introduction without a standing action. Gemini's no-SSA response is included for comparison.}
\label{tab:weak-ssa-case}
\end{table*}

\section{Judge Robustness}
\label{app:judge-robustness}

We assess judge robustness using three additional judges: DeepSeek-V4-Flash~\citep{deepseekai2026deepseekv4}, Gemini-3.8-Flash, and Kimi-K2.6\footnote{\url{https://www.kimi.ai/blog/kimi-k2-6}} in non-thinking mode. The SSA and SAR subsets cover 25\% of SAEIs (randomly selected). We compare model-pair score differences across judges.

% {\color{orange}For each task, we apply the main evaluation's pooling method to the same instances across judges and models, and compare model-pair score differences. SAR scores include length deductions.}

\begin{figure*}[t]
\centering
\includegraphics[width=\textwidth]{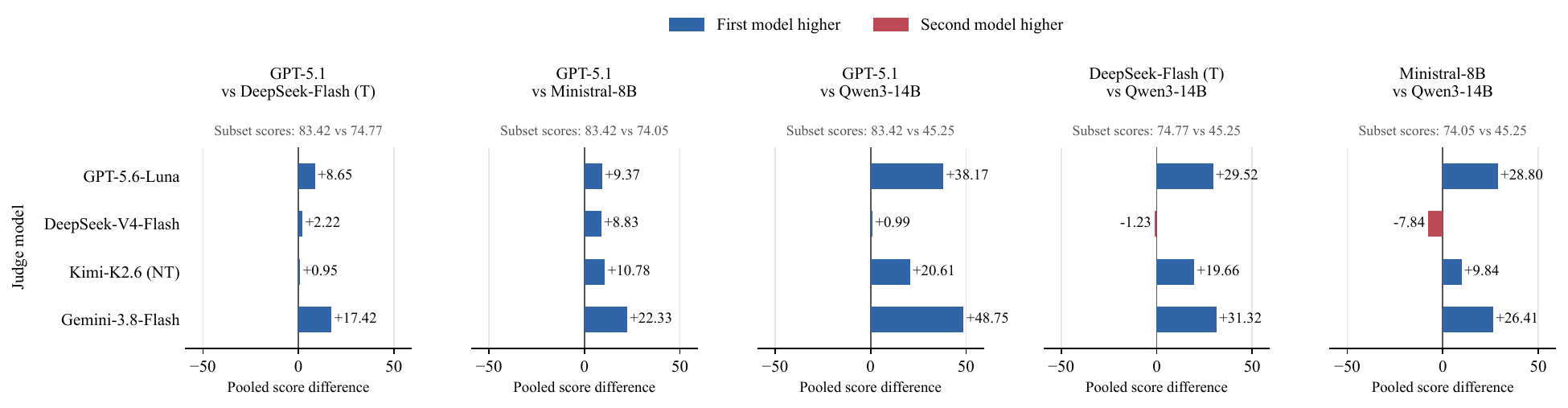}
\caption{Pooled SSA score differences under four judges. Positive values favor the first model; negative values favor the second. Subset scores are reported by GPT-5.6-Luna.}
\label{fig:judge-robustness}
\par\vspace{8pt}
\includegraphics[width=\textwidth]{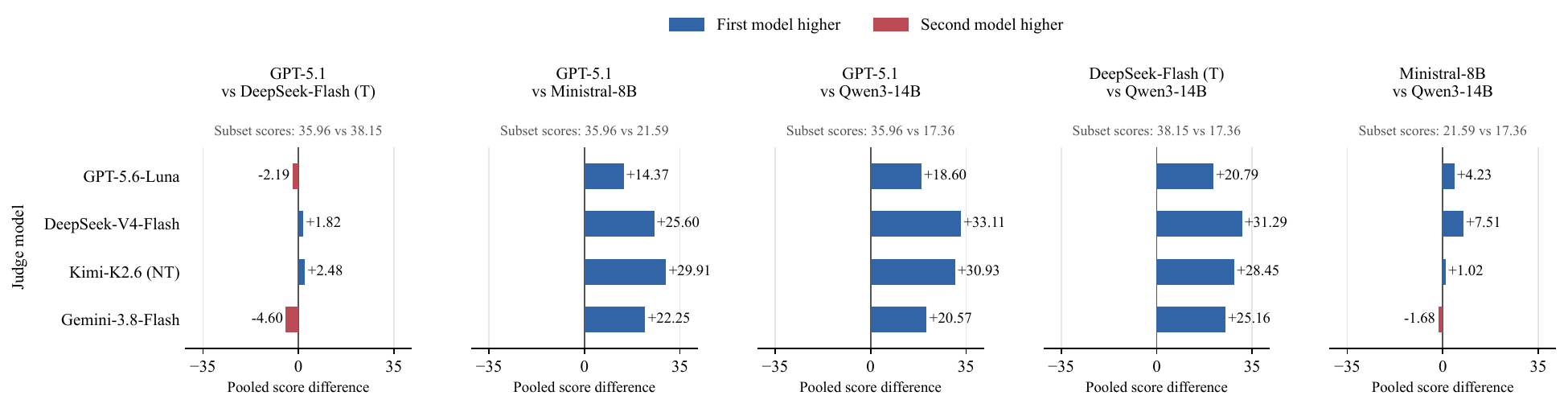}
\caption{Pooled SAR score differences under four judges, including length deductions. Positive values favor the first model; negative values favor the second. Subset scores are reported by GPT-5.6-Luna.}
\label{fig:sar-judge-robustness}
\end{figure*}

Figures~\ref{fig:judge-robustness} and~\ref{fig:sar-judge-robustness} show that most selected model pairs retain the same pooled-score ordering across judges. All four judges score GPT-5.1 above Ministral-8B and Qwen3-14B in both tasks. At least three judges also score DeepSeek-V4-Flash (T) and Ministral-8B above Qwen3-14B in each task. These agreements across judges from different model families largely support the robustness of our evaluation results. In SAR, GPT-5.6-Luna scores DeepSeek-V4-Flash (T) above GPT-5.1, while the DeepSeek judge gives the opposite ordering, providing evidence against a consistent own-family preference in these comparisons.

\subsection{Human--Judge Agreement}
\label{app:human-judge-agreement}
\begin{figure*}[t]
\centering
\includegraphics[width=\textwidth]{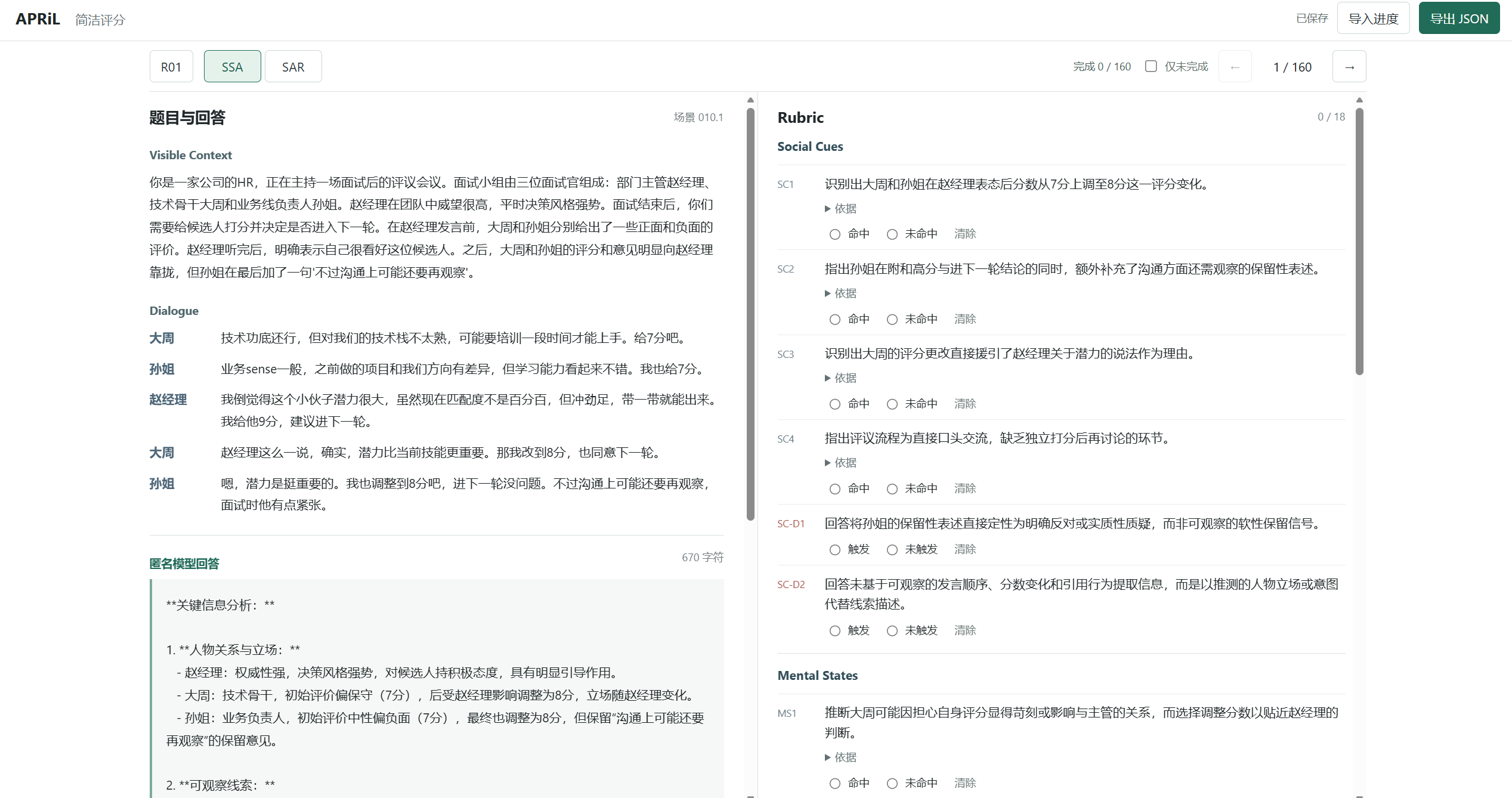}
\caption{Interface for human judging of model outputs.}
\label{fig:human-judge-interface}
\end{figure*}

\begin{figure*}[t]
\centering
\includegraphics[width=\textwidth]{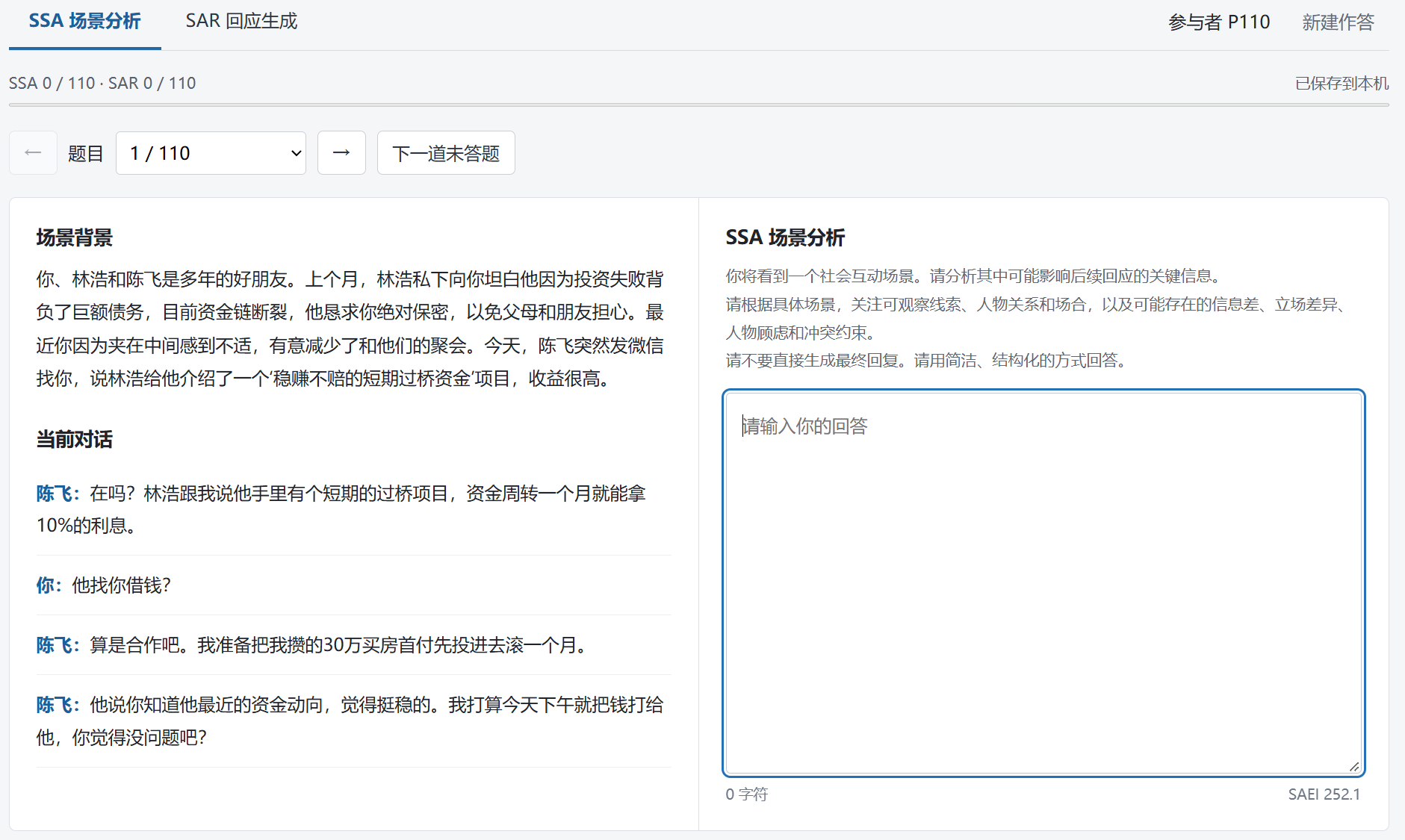}
\caption{Interface for human participants completing SSA and SAR.}
\label{fig:human-performance-interface}
\end{figure*}

Figures~\ref{fig:human-judge-interface} and~\ref{fig:human-performance-interface} show the interfaces for human judging and human task responses, respectively.

Three evaluators independently annotate the same model outputs from 40 scenarios. Each scenario includes outputs from GPT-5.1, DeepSeek-V4-Flash (T), Ministral-8B, and Qwen3-14B. Evaluators judge each rubric score point and deduction point without seeing model identities or LLM judgments. Fleiss' $\kappa$ is 0.424 / 0.413 for SSA and 0.596 / 0.515 for SAR (score points / deduction points), indicating moderate agreement.

Table~\ref{tab:human-judge-agreement} reports each judge's correlation with mean human scores on individual outputs and its difference from the pooled human score for each model. Luna correlates with human ratings in both tasks ($\rho=0.798$ for SSA and $0.874$ for SAR), achieves the highest SAR correlation, and has the smallest mean absolute score differences (4.64 and 2.54 points). Gemini achieves a higher SSA correlation but larger score differences. These results support Luna's selection as the main judge.

\begin{table*}[t]
\centering
\small
\setlength{\tabcolsep}{5pt}
\rowcolors{2}{white}{tablerowcolor}
\begin{tabularx}{\textwidth}{@{}l*{5}{>{\centering\arraybackslash}X}@{}}
\toprule
\rowcolor{tableheadcolor}
\textbf{Evaluated Model} & \textbf{Human} & \textbf{Luna} $\Delta$ & \textbf{DeepSeek} $\Delta$ & \textbf{Kimi} $\Delta$ & \textbf{Gemini} $\Delta$ \\
\midrule
\multicolumn{6}{l}{\textbf{SSA}} \\
GPT-5.1 & 88.68 & 1.02 & 3.94 & 4.64 & 21.40 \\
DeepSeek-V4-Flash (T) & 77.05 & 3.68 & 7.71 & 12.80 & 29.62 \\
Ministral-8B & 71.59 & 0.71 & 7.37 & 1.16 & 28.03 \\
Qwen3-14B & 57.81 & 13.16 & 37.41 & 12.38 & 39.74 \\
\cmidrule(lr){1-6}
Mean $\Delta\downarrow$ & -- & \textbf{4.64} & 14.11 & 7.74 & 29.70 \\
Spearman's $\rho\uparrow$ & -- & 0.798 & 0.176 & 0.654 & \textbf{0.869} \\
\midrule
\multicolumn{6}{l}{\textbf{SAR}} \\
GPT-5.1 & 34.78 & 2.17 & 15.22 & 10.87 & 17.39 \\
DeepSeek-V4-Flash (T) & 64.49 & 3.62 & 3.62 & 18.84 & 12.32 \\
Ministral-8B & 21.01 & 0.72 & 11.59 & 21.01 & 5.80 \\
Qwen3-14B & 15.94 & 3.62 & 0.72 & 15.94 & 1.45 \\
\cmidrule(lr){1-6}
Mean $\Delta\downarrow$ & -- & \textbf{2.54} & 7.79 & 16.67 & 9.24 \\
Spearman's $\rho\uparrow$ & -- & \textbf{0.874} & 0.768 & 0.845 & 0.809 \\
\bottomrule
\end{tabularx}
\caption{Agreement with human ratings. Human reports the mean pooled score of three evaluators; $\Delta$ is the absolute difference between each judge's pooled score and the human mean. Mean $\Delta$ averages these differences across four evaluated models; $\rho$ is the output-level correlation with mean human scores. Bold marks the smallest mean difference or highest correlation.}
\label{tab:human-judge-agreement}
\end{table*}

\section{Human Performance}
\label{app:human-performance}

\subsection{Results on the Same Subset}
The four human participants served as task respondents rather than scoring annotators. We randomly sampled 110 SAEIs from APRiL and compare model and human performance on this subset. Every participant completed both SSA and SAR for all 110 instances. Their answers were evaluated using the same LLM-as-a-Judge and rubrics as the model outputs. Because the participants generated analyses and responses rather than quantitative labels, we do not treat conventional inter-annotator agreement as an evaluation metric. We report the mean judge score across the four participants.

\begin{table*}[!htbp]
\centering
\small
\rowcolors{2}{white}{blue!5}
\resizebox{\textwidth}{!}{%
\begin{tabular}{lrrrrr}
\toprule
\rowcolor{blue!15} \textbf{Model} & \textbf{Social Cues} & \textbf{Mental States} & \textbf{Conflict Structure} & \textbf{SSA} & \textbf{SAR} \\
\midrule
GPT-5.1 & \underline{86.55} & \underline{75.11} & \underline{89.57} & \underline{83.74} & 39.43 \\
DS-V4-Pro (T) & \textbf{87.55} & \textbf{84.16} & \textbf{94.08} & \textbf{88.60} & 43.24 \\
DS-V4-Pro (NT) & 84.34 & 69.91 & 88.39 & 80.88 & 30.19 \\
Kimi-K2.5 (T) & 81.33 & 61.09 & 88.86 & 77.09 & \underline{44.91} \\
DS-V4-Flash (T) & 84.94 & 63.57 & 82.46 & 76.99 & 42.83 \\
Gemini-3-Pro & 81.12 & 61.54 & 85.07 & 75.91 & 43.96 \\
Gemini-3.8-Flash & 82.93 & 65.38 & 89.10 & 79.14 & \textbf{51.13} \\
Kimi-K2.5 (NT) & 84.54 & 56.11 & 80.81 & 73.82 & 36.79 \\
DS-V4-Flash (NT) & 79.92 & 56.56 & 78.20 & 71.56 & 32.08 \\
\specialrule{0.6pt}{0pt}{0pt}
Ministral-14B & \underline{79.92} & \underline{61.31} & \textbf{86.97} & \textbf{76.07} & \underline{21.32} \\
Ministral-8B & \textbf{81.12} & \textbf{63.12} & \underline{82.94} & \underline{75.73} & \textbf{22.08} \\
Ministral-3B & 64.46 & 41.63 & 68.01 & 58.03 & 7.55 \\
Qwen3-14B & 51.20 & 31.22 & 52.37 & 44.93 & 18.49 \\
Qwen3-8B & 42.57 & 24.66 & 41.94 & 36.39 & 16.60 \\
Qwen3-4B & 47.99 & 20.81 & 45.73 & 38.18 & 15.09 \\
\specialrule{0.6pt}{0pt}{0pt}
\textit{Black-box Avg} & \textit{83.69} & \textit{65.94} & \textit{86.28} & \textit{78.64} & \textit{40.51} \\
\textit{White-box Avg} & \textit{61.21} & \textit{40.46} & \textit{62.99} & \textit{54.89} & \textit{16.86} \\
\textit{LLM Avg} & \textit{74.70} & \textit{55.75} & \textit{76.97} & \textit{69.14} & \textit{31.05} \\
Human & 58.22 & 42.35 & 44.81 & 48.46 & 44.17 \\
\bottomrule
\end{tabular}%
}
\caption{Model and human performance on the same 110-instance subset. DS denotes DeepSeek; T and NT denote thinking and non-thinking modes, respectively. Bold and underlined values denote the best and second-best results within each model group.}
\label{tab:human-results}
\end{table*}

Table~\ref{tab:human-results} reports model and human performance on the shared subset. Human SSA is 30.18 points lower than the black-box average and 6.43 points lower than the open-weight average, leaving humans above only the three Qwen3 models. Mental States is also the lowest-scoring SSA dimension for human participants (42.35, compared with 58.22 for Social Cues and 44.81 for Conflict Structure). On SAR, humans exceed the black-box average by 3.66 points and the open-weight average by 27.31 points, although Gemini-3.8-Flash and Kimi-K2.5 (T) score higher. Humans thus show lower analysis scores alongside stronger response performance than both model-group averages.

Humans score lower on SSA but above both model-group averages on SAR. They may understand some social information without writing it down in their SSA responses, which may partly explain their lower SSA scores. Since SSA awards credit for explicitly stated points, analyses that state fewer relevant points may receive lower scores even when participants can produce appropriate responses. Participants may also rely more on intuition when responding. Their limited personal experience may make it difficult to empathize with people in unfamiliar situations or understand perspectives associated with roles they have not occupied, potentially affecting both their analyses and responses.

\subsection{Human Response Example}
The original Chinese human response and its English translation are presented in Table~\ref{tab:human-response-example-en}.
\begin{table*}[ht]
\centering
\small
\begin{tcolorbox}[appendixtext]
\appendixheading{Chinese Original}
\appendixfield{\kailabel{场景：}}
一名29岁的独生女在一线城市工作。父母长期催婚，并希望她接受一名在老家工作且经济条件较好的相亲对象。女儿因缺乏感情基础和职业规划不同而拒绝后，母亲强调父母多年的养育投入，并以父亲曾经生病住院却未告知女儿为例，要求她“听我们一次”。\par
\appendixfield{\kailabel{人类SSA回答：}}
“母亲嘴上说是为了自己好，实际上可能在完成自己作为父母的‘指标’，这忽略了女儿的个人感受，而且对方也在老家，两个人的职业发展也是不一样的。而且母亲可能存在自己用家长的身份来施压的可能性。以及用愧疚教育，如‘去年你爸生病住院，我都没告诉你，怕影响你工作。我们为你付出这么多，你就不能听我们一次？’给女儿带来压力。母亲说当年和父亲也是相亲认识，但实际上以往不同于现在，现在大家更希望走入自由恋爱而非被干涉。”\par
\appendixfield{\kailabel{评判结果：}}
\scoretwo{Social Cues、Mental States和Conflict Structure分别获得50、25和75分，SSA等权平均为50分。}该回答识别了父母身份、异地发展、养育投入和内疚压力，并将母亲可能把催婚视为履行父母责任作为一种解释。将住院经历理解为施加内疚压力具有题面依据，但若同时考虑母亲的焦虑、牺牲感或传统婚恋观，并进一步分析女儿的内疚、独立需求和家庭关系顾虑，场景分析会更加完整。\par
\appendixfield{\kailabel{人类SAR回答：}}
“妈，我知道你们肯定是爱我的，养我这么大也不容易，你们是我最亲的人，我不想让你们操心。我知道这件事，心里特别愧疚，以后你们别瞒我了。但是结婚这件事，是我一辈子的事，他在老家条件好我承认，但从我个人角度出发，我更希望在大城市发展，这是我个人的发展选择，而且我跟他也不合适，我们根本就不是一类人。我在这边也有我的生活。我希望你们能够尊重我的选择。你放心，我自己的事我心里有数，遇到合适的我不会拒绝的。我只是不想为了结婚而结婚。”\par
\appendixfield{\kailabel{评判结果：}}
\scorethree{SAR得分为75。}该回答回应了父母的关心与付出，同时说明了个人职业规划、关系判断和婚姻边界。其不足在于多次强调内疚，可能进一步强化父母通过付出施压的互动方式，且对后续如何继续沟通缺少更具体的安排。\par

\vspace{0.45em}
\appendixheading{English Translation}
\appendixfield{\textbf{Scenario:}}
A 29-year-old woman, an only child, works in a major city. Her parents have long pressured her to marry and want her to accept a well-off man who works in their hometown. She refuses because they lack an emotional connection and have different career plans. Her mother then stresses the years spent raising her and mentions that her father was once hospitalized without her being told, asking her to “listen to us just this once.”\par
\appendixfield{\textbf{Human SSA response:}}
“The mother says this is for her daughter's own good, but she may actually be trying to meet a parental ‘target.’ This overlooks her daughter's feelings. The man also lives in their hometown, so their career paths differ. The mother may be using her parental authority to apply pressure. She also uses guilt, saying, ‘When your dad was hospitalized last year, I did not tell you because I did not want to affect your work. We have done so much for you. Can you not listen to us just this once?’ This puts pressure on her daughter. The mother says she and the father also met through matchmaking, but times have changed. People today are more likely to want to choose their own partners without interference.”\par
\appendixfield{\textbf{Judgment:}}
\scoretwo{Social Cues, Mental States, and Conflict Structure receive 50, 25, and 75, giving an equally weighted SSA of 50.} The answer identifies parental authority, different locations and career paths, the parents' investment in raising their daughter, and guilt pressure. It also suggests that the mother may see arranging the marriage as fulfilling a parental duty. Interpreting the hospitalization story as guilt pressure is supported by the scenario. The analysis would be more complete if it also considered the mother's anxiety, sense of sacrifice, or traditional views on marriage, and the daughter's guilt, need for independence, and concerns about family relationships.\par
\appendixfield{\textbf{Human SAR response:}}
“Mom, I know you love me, and raising me was not easy. You and Dad are my closest family, and I do not want you to worry about me. Hearing this makes me feel very guilty. Please do not keep things like this from me again. But marriage affects my whole life. I accept that he is well-off in our hometown, but I want to build my career in a big city. That is my choice for my future. He and I are not compatible; we are very different people. I also have my own life here. I hope you can respect my choice. Please do not worry. I know what I am doing, and I will not turn down someone who is right for me. I just do not want to marry for the sake of getting married.”\par
\appendixfield{\textbf{Judgment:}}
\scorethree{SAR receives 75.} The response acknowledges the parents' concern and sacrifices while explaining the daughter's career plans, view of the relationship, and boundaries around marriage. Its repeated emphasis on guilt may reinforce the parents' use of sacrifice to apply pressure. It also lacks a more concrete plan for continuing the discussion.\par
\end{tcolorbox}
\caption{A bilingual human SSA and SAR example with judge evaluations.}
\label{tab:human-response-example-en}
\end{table*}

\section{Output Length and Task Scores}
\label{app:output-length-scores}

\begin{figure*}[t]
\centering
\includegraphics[width=\textwidth]{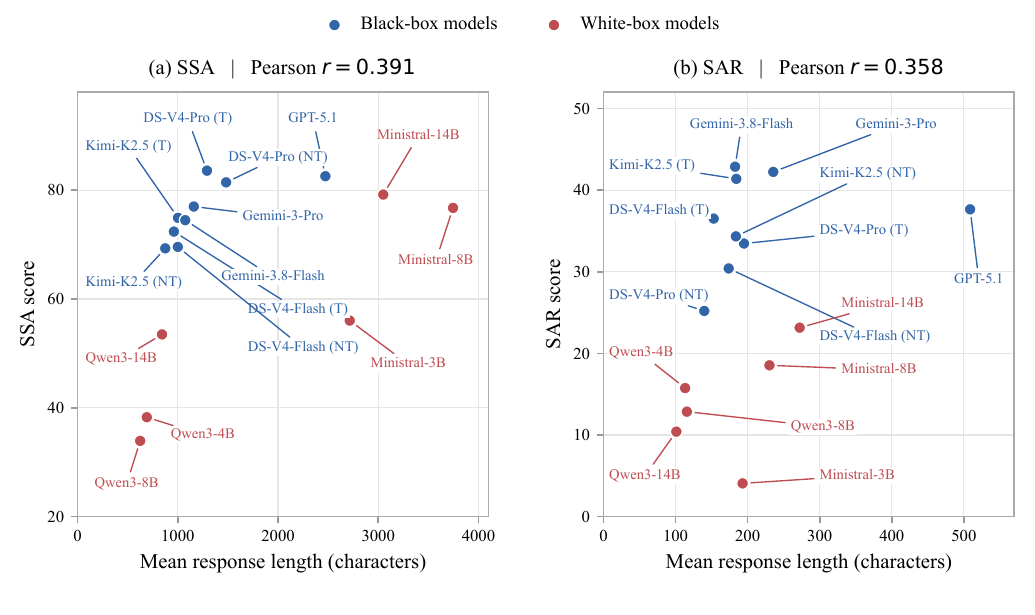}
\caption{Average output length and task scores. Each point represents a model and reasoning-mode setting. Colors distinguish black-box and white-box models. SAR scores include length deductions.}
\label{fig:output-length-scores}
\end{figure*}

We measure output length in characters, including punctuation, spaces, and line breaks, using responses with valid judge results for each task. Figure~\ref{fig:output-length-scores} compares average lengths and scores across 15 model and reasoning-mode settings. The Pearson correlations are 0.391 for SSA and 0.358 for SAR, indicating a weak-to-moderate positive relationship. This is plausible, as longer responses tend to cover more points and thus have more chances to match score points. However, length alone does not explain performance: for example, Ministral-8B produces the longest SSA outputs yet scores below GPT-5.1 and DeepSeek-V4-Pro (T). These correlations are computed across model settings rather than individual instances, and SAR scores include length deductions.

\end{CJK}

\end{document}